\documentclass{article} % For LaTeX2e
\usepackage{iclr2022_conference,times}
\iclrfinalcopy
\usepackage{amsmath,amsfonts,bm}

\def\eqref#1{equation~\ref{#1}}
\def\1{\bm{1}}

\DeclareMathAlphabet{\mathsfit}{\encodingdefault}{\sfdefault}{m}{sl}
\SetMathAlphabet{\mathsfit}{bold}{\encodingdefault}{\sfdefault}{bx}{n}

\usepackage{booktabs}       % professional-quality tables
\usepackage{amsfonts}       % blackboard math symbols
\usepackage{nicefrac}       % compact symbols for 1/2, etc.
\usepackage{microtype}      % microtypography
\usepackage{xcolor}         % colors
\usepackage{amsmath,amssymb} % define this before the line numbering.
\usepackage{hyperref}
\usepackage{url}
\usepackage{cleveref}
\usepackage{color}
\usepackage{multirow}
\usepackage{graphicx}
\usepackage{comment}
\usepackage{changepage}
\usepackage{wrapfig}
\usepackage{colortbl} 
\usepackage{arydshln} 
\newtheorem{lemma}{Lemma}

\newtheorem{assumption}{Assumption}
\newtheorem{proposition}{Proposition}

\usepackage{lipsum}
\newcommand{\tabincell}[2]{\begin{tabular}{@{}#1@{}}#2\end{tabular}}

\title{Memory Replay with Data Compression for Continual Learning}

\author{
Liyuan Wang\textsuperscript{\rm 1,2,3}\thanks{Equal Contribution} \And Xingxing Zhang\textsuperscript{\rm 1}\footnotemark[1] \And Kuo Yang\textsuperscript{\rm 6} \And Longhui Yu\textsuperscript{\rm 6} \And Chongxuan Li\textsuperscript{\rm 4,5}\thanks{Corresponding Authors} \\
\AND Lanqing Hong\textsuperscript{\rm 6} \And Shifeng Zhang\textsuperscript{\rm 6} \And Zhenguo Li\textsuperscript{\rm 6} \And Yi Zhong\textsuperscript{\rm 2,3}\footnotemark[2] \And Jun Zhu\textsuperscript{\rm 1}\footnotemark[2] \And \vspace{-.4cm}\\

\textsuperscript{\rm 1}Dept. of Comp. Sci. \& Tech., Institute for AI, BNRist Center, THBI Lab, Tsinghua University \\
\textsuperscript{\rm 2}School of Life Sciences, IDG/McGovern Institute for Brain Research, Tsinghua University \\
\textsuperscript{\rm 3}Tsinghua-Peking Center for Life Sciences
\textsuperscript{\rm 4}Gaoling School of AI, Renmin University of China\\ 
\textsuperscript{\rm 5}Beijing Key Laboratory of Big Data Management and Analysis Methods
\textsuperscript{\rm 6}Huawei Noah's Ark Lab\\
\texttt{\{wly19,xxzhang2020\}@mails.tsinghua.edu.cn,chongxuanli1991@gmail.com,}\\
\texttt{\{yang.kuo,honglanqing,zhangshifeng4,li.zhenguo\}@huawei.com,}\\
\texttt{yulonghui@stu.pku.edu.cn,\{zhongyithu, dcszj\}@tsinghua.edu.cn} \\

\vspace{-.4cm}
}

\begin{document}

\maketitle

\begin{abstract}
Continual learning needs to overcome catastrophic forgetting of the past. Memory replay of representative old training samples has been shown as an effective solution, and achieves the state-of-the-art (SOTA) performance. 
However, existing work is mainly built on a small memory buffer containing a few original data, which cannot fully characterize the old data distribution. In this work, we propose memory replay with data compression (MRDC) to reduce the storage cost of old training samples and thus increase their amount that can be stored in the memory buffer.
Observing that the trade-off between the quality and quantity of compressed data is highly nontrivial for the efficacy of memory replay, we propose a novel method based on determinantal point processes (DPPs) to efficiently determine an appropriate compression quality for currently-arrived training samples. 
In this way, using a naive data compression algorithm with a properly selected quality can largely boost recent strong baselines by saving more compressed data in a limited storage space.
We extensively validate this across several benchmarks of class-incremental learning and in a realistic scenario of object detection for autonomous driving.

\end{abstract}
\section{Introduction}

The ability to continually learn numerous tasks and infer them together is critical for deep neural networks (DNNs), which needs to mitigate \emph{catastrophic forgetting} \citep{mccloskey1989catastrophic} of the past. 
Memory replay of representative old training samples (referred to as \emph{memory replay}) has been shown as an effective solution, and achieves the state-of-the-art (SOTA) performance \citep{hou2019learning}.
Existing memory replay approaches are mainly built on a small memory buffer containing a few original data, and try to construct and exploit it more effectively. However, due to the low storage efficiency of saving original data, this strategy of building memory buffer will lose a lot of information about the old data distribution. On the other hand, this usually requires huge extra computation to further mitigate catastrophic forgetting, such as by learning additional parameters \citep{liu2021adaptive} or distilling old features \citep{hu2021distilling}.
%On the other hand, recent approaches in this direction attempted to improve the efficacy of memory replay by learning additional parameters \citep{Liu2020MANets} or distilling old features \citep{hu2021distilling}, which will largely increase the computation cost.
 
Different from ``artificial'' memory replay in DNNs, a significant feature of biological memory is to encode the old experiences in a highly compressed form and replay them to overcome catastrophic forgetting \citep{mcclelland2013incorporating, davidson2009hippocampal, carr2011hippocampal}. Thus the learned information can be maintained in a small storage space as comprehensively as possible, and flexibly retrieved. Inspired by the compression feature of biological memory replay, we propose memory replay with data compression (MRDC), which can largely increase the amount of old training samples that can be stored in the memory buffer by reducing their storage cost in a computationally efficient way. 
%As the operation of data compression only needs a small amount of extra computation, this strategy can both more effectively and more efficiently recover the old data distribution.

\begin{figure}[t]
    \centering
    \vspace{-0.4cm}
    \includegraphics[width=0.95\linewidth]{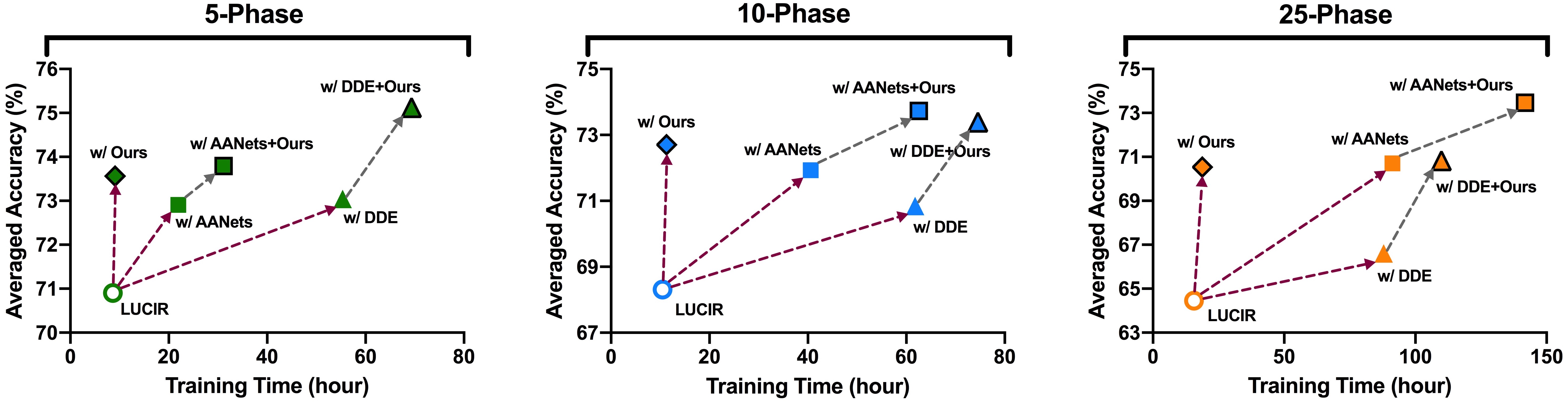}
    \vspace{-0.3cm}
    %\caption{Averaged accuracy and computation cost on ImageNet-sub. Compared with the SOTA approaches, memory replay with data compression (ours) achieves the SOTA performance with the lowest extra computation cost (purple arrow), and can further boost their performance (gray arrow). } 
        \caption{Averaged incremental accuracy and training time on ImageNet-sub. Using JPEG for data compression can achieve comparable or better performance than recent strong approaches with less extra computation (purple arrow), and can further improve their performance (gray arrow).} 
    \label{Time_Accuracy_ImageNet-sub_Summary}
    \vspace{-0.5cm}
\end{figure}

%However, it is often computation-inefficient to determine an appropriate quality of data compression by making a grid search, similarly as selecting other hyperparameters in continual learning. 
%However, it is often computation-inefficient to empirically determine the compression quality. \footnote{A naive grid search approach is to train continual learning processes with different qualities and choose the best one, resulting in huge computation cost.} 

Given a limited storage space, data compression introduces an additional degree of freedom to explicitly balance the quality and quantity for memory replay. 
%With a properly selected quality, using a naive JPEG algorithm \citep{wallace1992jpeg} of data compression can achieve the SOTA performance of class-incremental learning with the lowest extra computation (Fig. \ref{Time_Accuracy_ImageNet-sub_Summary}, purple arrow) and can further boost the SOTA memory replay approaches (Fig. \ref{Time_Accuracy_ImageNet-sub_Summary}, gray arrow). 
With a properly selected quality, using a naive JPEG compression algorithm \citep{wallace1992jpeg} can achieve comparable or better performance than recent strong approaches with less extra computation (Fig. \ref{Time_Accuracy_ImageNet-sub_Summary}, purple arrow), and can further improve their performance (Fig. \ref{Time_Accuracy_ImageNet-sub_Summary}, gray arrow).
However, to empirically determine the compression quality is usually inefficient and impractical, since it requires learning a task sequence or sub-sequence repeatedly\footnote{A naive grid search approach is to train continual learning processes with different qualities and choose the best one, resulting in huge computational cost. Also, this strategy will be less applicable if the old data cannot be revisited, or the future data cannot be accessed immediately.}. We propose a novel method based on determinantal point processes (DPPs) to efficiently determine it without repetitive training. Further, we demonstrate the advantages of our proposal in realistic applications such as continual learning of object detection for autonomous driving, where the incremental data are extremely large-scale. % with high storage cost.

Our contributions include: (1) We propose memory replay with data compression, which is both an important baseline and a promising direction for continual learning; (2) We empirically validate that the trade-off between quality and quantity of compressed data is highly nontrivial for memory replay, and provide a novel method to efficiently determine it without repetitive training;
%to efficiently determine an appropriate quality of data compression; 
%(3) Extensive experiments show that using a naive technique of data compression with a properly selected quality can substantially boost the efficacy of memory replay, and achieves the SOTA performance in a computation-efficient and plug-and-play way.
(3) Extensive experiments show that using a naive data compression algorithm with a properly selected quality can largely improve memory replay by saving more compressed data in a limited storage space.

\section{Related Work}
\textbf{Continual learning} needs to overcome catastrophic forgetting of the past when learning a new task. Regularization-based methods \citep{kirkpatrick2017overcoming, wang2021afec} approximated the importance of each parameter to the old tasks and selectively penalized its changes. Architecture-based methods \citep{rusu2016progressive} allocated a dedicated parameter subspace for each task to prevent mutual interference. Replay-based methods \citep{rebuffi2017icarl, shin2017continual} approximated and recovered the old data distribution.
%by saving a memory buffer or learning a generative model. 
In particular, memory replay of representative old training samples (referred to as \emph{memory replay}) can generally achieve the best performance in class-incremental learning \citep{liu2021adaptive,hu2021distilling} and in numerous other continual learning scenarios, such as audio tasks \citep{ehret2020continual}, few-shot \citep{tao2020few}, semi-supervised \citep{wang2021ordisco}, and unsupervised continual learning \citep{khare2021unsupervised}.
%In particular, it has been shown that remembering the old data distributions is necessary for class-incremental learning \citep{lesort2019regularization}, where memory replay of the old training samples achieves the SOTA performance. 
 
%Existing works in memory replay are mainly built on a small memory buffer containing a few original data, and attempt to construct and exploit it more effectively. 
Most of the work in memory replay attempted to more effectively construct and exploit a small memory buffer containing a few original data. As the pioneer work,  iCaRL \citep{rebuffi2017icarl} proposed a general protocol of memory replay for continual learning. To better construct the memory buffer, Mnemonics \citep{liu2020mnemonics} parameterized the original data and made them optimizable, while TPCIL \citep{tao2020topology} constructed an elastic Hebbian graph by competitive Hebbian learning. On the other hand, BiC \citep{wu2019large}, LUCIR \citep{hou2019learning}, PODNet \citep{douillard2020podnet}, DDE \citep{hu2021distilling} and AANets \citep{liu2021adaptive} attempted to better exploit the memory buffer, such as by mitigating the data imbalance between old and new classes \citep{hou2019learning,wu2019large,hu2021distilling}.

In contrast to saving original data, several work attempted to improve the efficiency of remembering the old data distribution.
%numerous efforts tried to remember the old data distribution more efficiently. 
One solution is to continually learn a generative model to replay generated data \citep{shin2017continual,wu2018memory} or compress old training data \citep{caccia2020online}. However, continual learning of such a generative model is extremely challenging, which limits its applications to relatively simple domains, and usually requires a lot of extra computation. Another solution is feature replay: GFR \citep{liu2020generative} learned a feature generator from a feature extractor to replay generated features, but the feature extractor suffered from catastrophic forgetting since it was incrementally updated. REMIND \citep{hayes2020remind} saved the old features and reconstructed the synthesized features for replay, but it froze the majority of feature extractor after learning the initial phase, limiting the learning of representations for incremental tasks.

\textbf{Data compression} aims to improve the storage efficiency of a file, including lossless compression and lossy compression. Lossless compression needs to perfectly reconstruct the original data from the compressed data, which limits its compression rate \citep{shannon1948mathematical}. In contrast, lossy compression can achieve a much higher compression rate by degrading the original data, so it has been broadly used in realistic applications. Representative hand-crafted approaches include JPEG (or JPG) \citep{wallace1992jpeg}, which is the most commonly-used algorithm of lossy compression \citep{mentzer2020high}, WebP \citep{lian2012webp} and JPEG2000 \citep{rabbani2002jpeg2000}. On the other hand, neural compression approaches generally rely on optimizing Shannon's rate-distortion trade-off, through RNNs \citep{toderici2015variable,toderici2017full}, auto-encoders \citep{agustsson2017soft} and GANs \citep{mentzer2020high}.

\section{Continual Learning Preliminaries}
We consider a general setting of continual learning that a deep neural network (DNN) incrementally learns numerous tasks from their task-specific training dataset \(D_{t} = \{(x_{t,i}, y_{t,i})\}_{i=1}^{{N}_{t}} \), where \(D_{t}\) is only available when learning task \(t\), \((x_{t,i}, y_{t,i})\) is a data-label pair and \(N_t\) is the number of such training samples. For classification tasks, the training samples of each task might be from one or several new classes. All the classes ever seen are evaluated at test time, and the classes from different tasks need to be distinguished. This setting is also called class-incremental learning \citep{vandeven2019three}. Suppose such a DNN with parameter \(\theta\) has learned \(T\) tasks and attempts to learn a new task. Since the old training datasets \(\bigcup_{t=1}^{T} D_{t}\) are unavailable, the DNN will adapt the learned parameters to fit \(D_{T+1}\), and tend to catastrophically forget the old tasks \cite{mcclelland1995there}.

%We consider a general setting of continual learning, which needs to incrementally learn numerous tasks $t = 1, 2, 3, ...$ from their task-specific training dataset \(D_{t} = \{(x_{t,i}, y_{t,i})\}_{i=1}^{{N}_{t}} \), where \((x_{t,i}, y_{t,i})\) is a data-label pair and \(N_t\) is the number of such training samples. For classification tasks, the training samples of each task might be from one or several new classes, i.e., class-incremental learning. For continual learning, \(D_{t}\) is only available when learning task \(t\), and all the tasks ever seen are evaluated at test time, which is known as the single-head evaluation \citep{chaudhry2018riemannian}. Suppose a DNN with parameter \(\theta\) has incrementally learned \(T\) tasks. When learning a new task, since the old training datasets \(\bigcup_{t=1}^{T} D_{t}\) are unavailable, the DNN will adapt the learned parameters to fit \(D_{T+1}\), and tend to catastrophically forget the old tasks \citep{mccloskey1989catastrophic}.

An effective solution of overcoming catastrophic forgetting is to select and store representative old training samples \(D_{t}^{mb} = \{(x_{t,i}, y_{t,i})\}_{i=1}^{{N}_{t}^{mb}} \) in a small memory buffer (\(mb\)), and replay them when learning the new task. For classification tasks, mean-of-feature is a widely-used strategy to select \(D_{t}^{mb}\) \citep{rebuffi2017icarl, hou2019learning}. 
%\citep{rebuffi2017icarl, riemer2018learning, hou2019learning, douillard2020podnet,Liu2020MANets,hu2021distilling}. 
After learning each task, features of the training data can be obtained by the learned embedding function \(F_{\theta}^{e}(\cdot)\). In each class, several data points nearest to the mean-of-feature are selected into the memory buffer. Then, the training dataset of task \(T+1\) becomes \(D'_{T+1} = D_{T+1} \bigcup {D}_{1:T}^{mb} \), including both new training samples \(D_{T+1} \) and some old training samples \({D}_{1:T}^{mb} = \bigcup_{t=1}^{T} {D}_{t}^{mb}\), so as to prevent forgetting of the old tasks.

However, due to the limited storage space, only a few original data can be saved in the memory buffer, namely, \({N}_{t}^{mb} \ll {N}_{t}\). Although numerous efforts in memory replay attempted to more effectively exploit the memory buffer, such as by alleviating the data imbalance between the old and new classes, this strategy of building memory buffer is less effective for remembering the old data distribution. %Several recent approaches attempted to improve the efficacy of memory replay by learning additional parameters \citep{liu2021adaptive} or distilling old features \citep{hu2021distilling}, but resulting in huge computation cost.

\section{Method}
In this section, we first present memory replay with data compression for continual learning. Then, we empirically validate that there is a trade-off between the quality and quantity of compressed data, which is highly nontrivial for memory replay, and propose a novel method to determine it efficiently.

\subsection{Memory Replay with Data Compression }

Inspired by the biological memory replay that is in a highly compressed form \citep{carr2011hippocampal},
%\citep{dragoi2006temporal, davidson2009hippocampal, carr2011hippocampal},
we propose an important baseline for memory replay, that is, using data compression to increase the amount of old training samples that can be stored in the memory buffer, so as to more effectively recover the old data distribution. Data compression can be generally defined as a function \(F_{q}^c(\cdot)\) of compressing the original data \(x_{t,i}\) to \(x_{q, t,i} = F_{q}^c(x_{t,i}) \) with a controllable quality \(q\). Due to the smaller storage cost of each \(x_{q, t,i}\) than \(x_{t,i}\), the memory buffer can maintain more old training samples for replay, namely, \(N_{q, t}^{mb} > {N}_{t}^{mb}\) for \(D_{q, t}^{mb} = \{(x_{q,t,i}, y_{t,i})\}_{i=1}^{N_{q,t}^{mb}} \) in \(D_{q, 1:T}^{mb} = \bigcup_{t=1}^{T} D_{q, t}^{mb}\). 
For notation clarity, we will use \({D}_{q}^{mb}\) to denote \({D}_{q, t}^{mb}\) without explicitly writing out its task label \(t\), likewise for \(x_{i}\), \(y_i\), \({N}_{q}^{mb}\), \({N}^{mb}\) and \({D}\). The compression rate can be defined as  \(r_{q} = N_{q}^{mb} / {N}^{mb} \propto N_{q}^{mb}\).
%Then, the compression rate can be defined as  \(r_{c} = N_{q, t}^{mb} / {N}_{t}^{mb} \propto N_{q, t}^{mb}\), depending on the quality \(q\).
%Since the compression rate of lossy compression is generally much higher than lossless compression \citep{shannon1948mathematical}, we mainly consider lossy compression for memory replay. This leads to an intuitive trade-off between quality and quantity: Due to the limited storage space, reducing the quality for a larger compression rate will increase the quantity of compressed data that can be stored in the memory buffer, and vice versa.

\begin{figure}[t]
%\begin{wrapfigure}{r}{0.50\textwidth}
    \centering
     \vspace{-0.3cm}
    \includegraphics[width=1\linewidth]{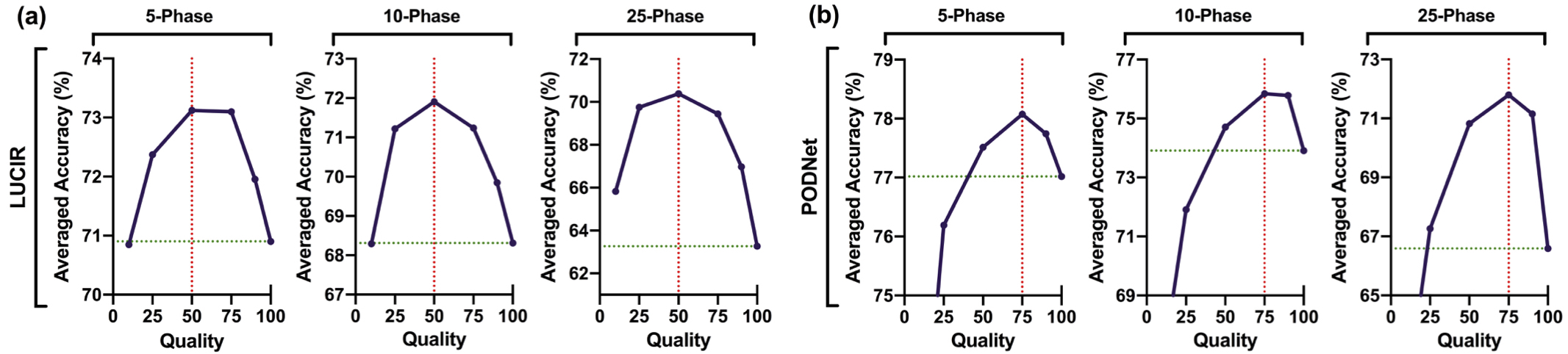}
     \vspace{-0.6cm}
    \caption{Memory replay with data compression on ImageNet-sub. We make a grid search of the JPEG quality in \(\{10, 25, 50, 75, 90\}\). The quality of 100 refers to the original data without compression.}      
%Using the same storage space as \(20\) original images per class, we make a grid search of the quality of JPEG among \(10\), \(25\), \(50\), \(75\) and \(90\), with 200, 125, 85, 58 and 37 compressed images per class, respectively. We present the results of 5-, 10- and 25-phase with LUCIR and PODNet. 
     \vspace{-0.5cm}
    \label{Accuracy_Tradeoff}
    %\vspace{-0.2cm}
%\end{wrapfigure}
\end{figure}

In analogy to the learning theory for supervised learning, we argue that continual learning will also benefit from replaying more compressed data, assuming that they approximately follow the original data distribution. However, the assumption is likely to be violated if the compression rate is too high. Intuitively, this leads to a trade-off between quality and quantity: if the storage space is limited, reducing the quality \(q\) of data compression will increase the quantity \(N_{q}^{mb}\) of compressed data that can be stored in the memory buffer, and vice versa.

%Although lossless compression can perfectly reconstruct original data from compressed data, its compression rate is generally much lower than that of lossy compression \citep{shannon1948mathematical}, so we mainly consider lossy compression for memory replay.
%Due to the limited compression rate of lossless compression \citep{shannon1948mathematical}, we mainly consider lossy compression to improve memory replay.
Here we evaluate the proposed idea by compressing images with JPEG \citep{wallace1992jpeg}, a naive but commonly-used lossy compression algorithm. JPEG can save images with a quality in the range of \([1, 100]\), where reducing the quality results in a smaller file size. Using a memory buffer equivalent to 20 original images per class \citep{hou2019learning}, we make a grid search of the JPEG quality with representative memory replay approaches, such as LUCIR \citep{hou2019learning} and PODNet \citep{douillard2020podnet}.
As shown in Fig. \ref{Accuracy_Tradeoff}, memory replay of compressed data with an appropriate quality can substantially outperform that of original data. However, whether the quality is too large or too small, it will affect the performance. In particular, the quality that achieves the best performance varies with the memory replay approaches, but is consistent for different numbers of splits of incremental phases.

%To more explicitly show how the quality-quantity trade-off affects memory replay, we use t-SNE \citep{van2008visualizing} to visualize the normalized features of compressed data and the same amount of original data in Fig. \ref{Feature_Manifold_Compressed}. With the decrease of quality and the increase of quantity, the area of compressed data is initially similar to that of original data and expands synchronously. However, as a large number of low-quality compressed data occur out-of-distribution, the area of compressed data becomes much larger than that of original data, where the performance also severely declines (see Fig. \ref{Accuracy_Tradeoff}).
%Therefore, reducing the quality to increase the quantity, the compressed data tend to be distorted and thus become out-of-distribution. 

\subsection{Quality-Quantity Trade-off}

Since the trade-off between quality $q$ and quantity $N_q^{mb}$ is highly nontrivial for memory replay, it needs to be carefully determined. Formally, after learning each task from its training dataset \(D\), let's consider several compressed subsets \(D_{q}^{mb} = \{(x_{q, i}, y_{i})\}_{i=1}^{{N}_{q}^{mb}} \), where \(q\) is from a set of finite candidates \(Q= \{q_1, q_2, q_3, ...\}\).
Each \(D_{q}^{mb}\) is constructed by selecting a subset \({D}^{mb *}_q = \{({x}_{i}, y_{i})\}_{i=1}^{{N}_{q}^{mb}} \) of \({N}_{q}^{mb}\) original training samples from \(D\) (following mean-of-feature or other principles). The size \({N}_{q}^{mb}\) is determined as the maximum number such that the compressed version of \({D}^{mb *}_q\) to a quality \(q\) can be stored in the memory buffer (see details in Appendix A). Thereby, a smaller \(q\) enables to save a larger \({N}_{q}^{mb}\), and vice versa. The objective is to select a compressed subset that can best represent \(D\), namely, to determine an appropriate \(q \) for memory replay. 

%let's consider a union dataset \(U_Q =  \bigcup_{q \in Q} D_{q}^{mb}\) of several compressed subsets \(D_{q}^{mb} = \{(x_{q, i}, y_{i})\}_{i=1}^{{N}_{q}^{mb}} \), where \(Q= \{q_1, q_2, q_3, ...\}\) includes the candidate values of \(q\). 
%we need to determine an appropriate quality \(q\) from several candidate values \(q \in Q= \{q_1, q_2, q_3, ...\}\), to compress and save old data in a limited memory buffer space (\(mb\)). 
%The determined quality \(q\) should reflect the effects of data compression approaches \(F_{c}^{q}(\cdot)\), and memory replay approaches for continual learning of the embedding function \(F_{e}^{\theta}(\cdot)\), both of which are integrated to the feature embedding \(f_{q,i} = F_{e}^{\theta}(F_{c}^{q}(x_{i}))\). 

To understand the effects of data compression, which depends on the compression function \(F_{q}^{c}(\cdot)\) and the continually-learned embedding function \(F_{\theta}^{e}(\cdot)\), we focus on analyzing the features of compressed data \(f_{q,i} = F_{\theta}^{e}(F_{q}^{c}(x_{i}))\).
%The effects of data compression to memory replay depend on the data compression approaches \(F_{q}^{c}(\cdot)\), and the memory replay approaches for continual learning of the embedding function \(F_{\theta}^{e}(\cdot)\). 
%Since both of them are integrated to the feature embedding \(f_{q,i} = F_{\theta}^{e}(F_{q}^{c}(x_{i}))\), 
We first calculate the feature matrix \(M_q^c = [\bar{f}_{q, 1}, \bar{f}_{q, 2}, ..., \bar{f}_{q, N_q^{mb}}]\) of each compressed subset \(D_{q}^{mb}\), where each column vector \(\bar{f}_{q, i}\) is obtained by normalizing \(f_{q, i}\) under \(L_2\)-norm to keep \(||\bar{f}_{q, i}||_2 = 1\). Similarly, we obtain the feature matrix \(M_q^{*}= [\bar{f}_{1}, \bar{f}_{2}, ..., \bar{f}_{N_q^{mb}}]\) of each original subset \(D_q^{mb *}\). Then, we can analyze the quality-quantity trade-off from two aspects:

%Since the determined \(q \) should integrate the effects of data compression approaches \(F_{c}^{q}(\cdot)\), and memory replay approaches for continual learning of the embedding function \(F_{e}^{\theta}(\cdot)\), we analyze the quality-quantity trade-off with feature embedding of compressed data \(f_{q,i} = F_{e}^{\theta}(F_{c}^{q}(x_{i}))\). 
\begin{figure}[th]
    \centering
     \vspace{-0.2cm}
    \includegraphics[width=1\linewidth]{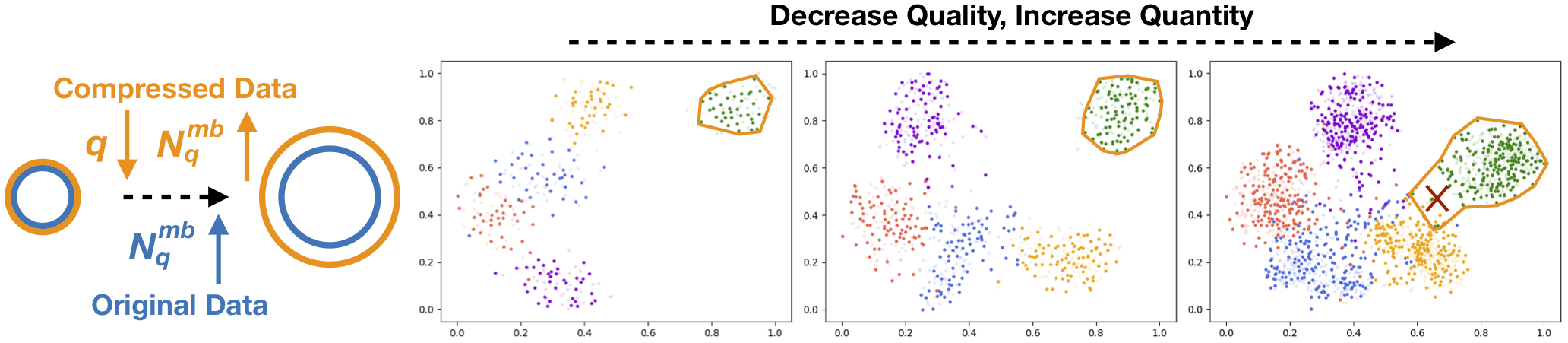}
    \vspace{-0.6cm}
    \caption{t-SNE visualization of features of the original subset (light dots) and its compressed subset (dark dots) after learning 5-phase ImageNet-sub with LUCIR. From left to right, the quantity is increased from 37, 85 to 200, while the JPEG quality is reduced from 90, 50 to 10. We plot five classes out of the latest task and label them in different colors. The crossed area is out-of-distribution. } 
%the JPEG quality is reduced from 90, 50 to 10 to increase the quantity from 37, 85 to 200. 
%Using the same storage space as \(20\) original images per class, the quality of compressed data in the three column is \(q = \{90, 50, 10\}\) while the quantity is  \(N_{q}^{mb} = \{37, 85, 200\}\) images per class, respectively. 
    \label{Feature_Manifold_Compressed}
    \vspace{-0.3cm}
\end{figure}

On the \textbf{empirical} side, in Fig. \ref{Feature_Manifold_Compressed} we use t-SNE \citep{van2008visualizing} to visualize features of the original subset (light dot), which includes different amounts of original data, and its compressed subset (dark dot), which is obtained by compressing the original subset to just fit in the memory buffer. With the increase of quantity and the decrease of quality, the area of compressed subset is initially similar to that of original subset and expands synchronously. However, as a large number of low-quality compressed data occur out-of-distribution, the area of compressed subset becomes much larger than that of its original subset, where the performance also severely declines (see Fig. \ref{Accuracy_Tradeoff}, \ref{Feature_Manifold_Compressed}).

\iffalse
On the \textbf{theoretical} side, given a training dataset $D$, we aim to find the compressed subset \(D_q^{mb}\) that can best represent \(D\) by choosing an appropriate compression quality \(q\) from its range.
To achieve this goal, we introduce $\mathcal{P}_{q} (D_q^{mb}|D)$ to characterize the conditional likelihood of selecting \(D_q^{mb}\) given input \(D\) under parameter $q$. Since the network parameter $\theta$ is fixed during compression, and the feature matrix $M_q^c=F_{\theta}^e(D_q^{mb})$, we rewrite $\mathcal{P}_{q} (D_q^{mb}|D)$ as $\mathcal{P}_{q} (M_q^c|D)$ equivalently. By focusing on maximum likelihood estimation (MLE), we propose to solve
\begin{equation}\label{Compress-obj} 
\begin{split}
  \max_{q} \ \mathcal{P}_{q} (M_q^c|D).
  \end{split}
\end{equation}
We draw inspirations from determinantal point processes (DPPs), which are elegant probabilistic sampling models in data selection fields \citep{kulesza2012determinantal}. 
In particular, DPPs specify the probabilities for every possible subset with the volume spanned by all elements in this subset, using determinants.
%In particular, DPPs specify the probabilities for every possible subset with the volume of the parallelepiped spanned by all elements in this subset, using determinants.
Depending on the nice properties of DPPs (e.g., high-quality diverse selection), our goal can be modeled as a DPP, i.e., a distribution over all subsets of $D$ with cardinality $N_q^{mb}$ (detailed in Appendix C.2).
Formally, such a DPP formulates the probability $\mathcal{P}_{q} (M_q^c|D)$ as
\fi

On the \textbf{theoretical} side, given a training dataset $D$, we aim to find the compressed subset \(D_q^{mb}\) that can best represent \(D\) by choosing an appropriate compression quality \(q\) from its range.
To achieve this goal, we introduce $\mathcal{P}_{q} (D_q^{mb}|D)$ to characterize the conditional likelihood of selecting \(D_q^{mb}\) given input \(D\) under parameter $q$. The goal of learning is to choose appropriate $q$ based on the training tasks for making accurate predictions on unseen inputs.
% Since the network parameter $\theta$ is fixed during compression, and the feature matrix $M_q^c=F_{\theta}^e(D_q^{mb})$, we rewrite $\mathcal{P}_{q} (D_q^{mb}|D)$ as $\mathcal{P}_{q} (M_q^c|D)$ equivalently.
While there are a variety of objective functions for learning, here we focus on the widely-used maximum likelihood estimation (MLE), where the goal is to choose $q$ to maximize the conditional likelihood of the observed data:
% % $\mathcal{L}(q)=\mathcal{P}_{q} (D^{mb}|D)$.
% % By focusing on maximum likelihood estimation (MLE), we propose to solve
\begin{equation}\label{Compress-obj} 
\begin{split}
  \max_{q} \ \mathcal{P}_{q} (D_q^{mb}|D).
  \end{split}
\end{equation}
The construction of $D_q^{mb}$ can be essentially viewed as a sampling problem with the cardinality ${N}_q^{mb}$.
Here, we apply Determinantal Point Processes (DPPs) to formulate the conditional likelihood $\mathcal{P}_{q} (D_q^{mb}|D)$, since DPPs are not only elegant probabilistic sampling models \citep{kulesza2012determinantal}, which can characterize the probabilities for every possible subset by determinants, but also provide a geometric interpretation of the probability by the volume spanned by all elements in the subset (detailed in Appendix C.2).
In particular, a conditional DPP is a conditional probabilistic model which assigns a probability $\mathcal{P}_{q} (D_q^{mb}|D)$ to each possible subset $D_q^{mb}$. 
Since the network parameter $\theta$ is fixed during compression, and the feature matrix $M_q^c=F_{\theta}^e(D_q^{mb})$, we rewrite $\mathcal{P}_{q} (D_q^{mb}|D)$ as $\mathcal{P}_{q} (M_q^c|D)$ equivalently.
%=\frac{\det(L_{D_q^{mb}}(D; q)}{\sum_{|{D'}|=N_q^{mb}}\det(L_{{D'}}(D; q))}
% We draw inspirations from determinantal point processes (DPPs), which are elegant probabilistic sampling models in data selection fields \citep{kulesza2012determinantal}. 
% In particular, DPPs specify the probabilities for every possible subset with the volume spanned by all elements in this subset, using determinants.
% %In particular, DPPs specify the probabilities for every possible subset with the volume of the parallelepiped spanned by all elements in this subset, using determinants.
% Depending on the nice properties of DPPs (e.g., high-quality diverse selection), our goal can be modeled as a DPP, i.e., a distribution over all subsets of $D$ with cardinality $N_q^{mb}$ (detailed in Appendix C.2).
Formally, such a DPP formulates the probability $\mathcal{P}_{q} (D_q^{mb}|D)$ as
\begin{equation}\label{Compress-def} 
\begin{split}
  \mathcal{P}_{q} (M_q^c|D) 
  = \frac{\det(L_{M_q^c}(D; q,\theta))}{\sum_{|{M}|=N_q^{mb}}\det(L_{{M}}(D; q, \theta))},
  \end{split}
\end{equation}
where $|{M_q^c}|=N_q^{mb}$ and \(L(D; q, \theta) \) is a conditional DPP $|D|\times |D|$ kernel matrix that depends on the input $D$, the parameters $\theta$ and $q$. 
$L_{M}(D; q, \theta)$ (resp., $L_{M_q^c}(D; q,\theta)$) is the submatrix sampled from \(L(D; q, \theta) \) using indices from $M$ (resp., $M_q^c$).
The numerator defines the marginal probability of inclusion for the subset ${M_q^c}$, and the denominator serves as a normalizer to enforce the sum of $\mathcal{P}_{q} (M_q^c|D)$ for every possible ${M_q^c}$ to 1. Generally, there are many ways to obtain a positive semi-definite kernel $L$. In this work, we employ the most widely-used dot product kernel function, where $L_{M_q^c}=M_q^{c \top} M_q^{c}$ and  $L_{{M}}={M}^{ \top} {M}$.

However, due to the extremely high complexity of calculating the denominator in Eq.~(\ref{Compress-def}) (analyzed in Appendix C.1), it is difficult to optimize \(\mathcal{P}_{q} (M_q^c|D)\). Alternatively, by introducing \(\mathcal{P}_{q} (M_q^*|D) \) to characterize the conditional likelihood of selecting \(M_q^*\) given input \(D\) under parameter $q$, we propose a relaxed optimization program of Eq.~(\ref{Compress-obj}), in which we (1) maximize \(\mathcal{P}_{q} (M_q^*|D) \) since \(\mathcal{P}_{q} (M_q^c|D) \le \mathcal{P}_{q} (M_q^*|D) \) is always satisfied under lossy compression; and meanwhile, (2) constrain that \(\mathcal{P}_{q} (M_q^c|D) \) is consistent with \(\mathcal{P}_{q} (M_q^*|D) \). The relaxed program is solved as follows.

First, by formulating $\mathcal{P}_{q} (M_q^*|D)$ similarly as $\mathcal{P}_{q} (M_q^c|D)$ in Eq.~(\ref{Compress-def}) (detailed in Appendix C.2), we need to maximize
\begin{equation}\label{Uncompess}
\begin{split}
  %\mathcal{L}_1(q) = \mathcal{P}_{q} (M_q^*|D) =  \frac{\det(L_{M_q^*}(D;\theta))}{\sum_{|M^*|=N_q^{mb}} \det(L_{M^*}(D;  \theta))}  \propto \det({M}_q^{* \top} M_q^{*}) = (\rm{Vol}_q^{*})^{2},
    \mathcal{L}_1(q) = \mathcal{P}_{q} (M_q^*|D) = 
  \frac{\det(L_{M_q^*}(D;\theta))}{\sum_{|M^*|=N_q^{mb}} \det(L_{M^*}(D;  \theta))},
  \end{split}
\end{equation} 
%where $|{M_q^*}|=N_q^{mb}$, \(L(D; \theta) \) is a conditional DPP $|D|\times |D|$ kernel matrix that depends on the input $D$ and parameters $\theta$.$L_{M^*}(D; \theta)$ (resp., $L_{M_q^*}(D; q,\theta))$) is the submatrix sampled from \(L(D; \theta) \) using indices from $M^*$ (resp., $M_q^*$). 
where the conditional DPP kernel matrix \(L(D; \theta) \) only depends on $D$ and $\theta$. For our task, $\mathcal{P}_{q} (M_q^*|D) $ monotonically increases with $N_q^{mb}$. Thus, optimizing $\mathcal{L}_1$ is converted into $\max_q\ {N_q^{mb}}$ equivalently, with significantly reduced complexity (detailed in Proposition~\ref{propo1} of Appendix C.3).

Second, to constrain that $\mathcal{P}_q (M_q^c|D)$ is consistent with $\mathcal{P}_q (M_q^{*}|D)$, we propose to minimize
\begin{equation}
\begin{split}
    \mathcal{L}_2(q)& =
     \left |  \frac{ \mathcal{P}_q (M_q^c|D) }{ \mathcal{P}_q (M_q^{*}|D)} -1 \right | 
     =  \left | \frac{\det(M_q^{c \top} M_q^c)}{\det(M_q^{* \top} M_q^{*})}Z_q  -1 \right |
    =  \left | \left(\frac{\rm{Vol}_q^c}{\rm{Vol}_q^{*}}\right)^2 Z_q -1 \right | 
    =  \left |  R_q^{2} Z_q -1 \right |,
     %\mathcal{L}_2(q)& =
     %\left |  \frac{ \mathcal{P}_q (M_q^c|D) }{ \mathcal{P}_q (M_q^{*}|D)} -1 \right | 
    %=  \left |  \frac{ \det(L_{M_q^c}(D; q,\theta)) }{\det(L_{M_q^{*}}(D;\theta))}Z_q -1 \right | \\
    %& =  \left | \frac{\det(M_q^{c \top} M_q^c)}{\det(M_q^{* \top} M_q^{*})}Z_q  -1 \right |
    %=  \left | \left(\frac{\rm{Vol}_q^c}{\rm{Vol}_q^{*}}\right)^2 Z_q -1 \right | 
    %=  \left |  R_q^{2} Z_q -1 \right |,
\end{split}
\end{equation}
where $Z_q=\frac{\sum_{|M^*|=N_q^{mb}} \det(L_{M^*}(D;  \theta))}{\sum_{|M|=N_q^{mb}} \det(L_{M}(D;  q, \theta))}$. In particular, $\det(M_q^{* \top} M_q^*)$ has a geometric interpretation that it is equal to the squared volume spanned by $M_q^*$ \citep{kulesza2012determinantal}, denoted as $\rm{Vol}_q^{*}$, likewise for $\det(M_q^{c \top} M_q^c)$ with respect to $\rm{Vol}_q^{c}$. Then we define \(R_q = \frac{\rm{Vol}_q^c}{\rm{Vol}_q^{*}}\) as the ratio of the two feature volumes. 
%$N_q^{mb}$-dimensional 
%In particular, $\det(M_q^{* \top} M_q^*)$ has a geometric interpretation that it is equal to the squared $N_q^{mb}$-dimensional volume of the parallelepiped spanned by $M_q^*$, denoted as $\rm{Vol}_q^{*}$  \citep{kulesza2012determinantal}, likewise for $\det(M_q^{c \top} M_q^c)$ and $\rm{Vol}_q^{c}$. 
% % In essence, we can convert the constraint on \(\mathcal{L}_2 \) into constraining \(|R_q - 1|< \epsilon \), 
% so as to avoid computing \(Z_q\) (detailed in Proposition 2 of Appendix C). The hyperparameter $\epsilon$ is a small positive number to serve as the threshold of \(R_q\). Since \(R_q\) is calculated by normalizing the feature volume $\rm{Vol}_q$ with $\rm{Vol}_q^{*}$, both of which depend on $q$ (i.e., $N_q^{mb}$), its threshold $\epsilon$ largely mitigates the sensitivity of $q$ to various domains and can be empirically chosen as a constant value.
To avoid computing \(Z_q\), we can convert optimizing \(\mathcal{L}_2 \) into minimizing \(|R_q - 1| \) equivalently, since both of them mean maximizing $q$ (detailed in Proposition~\ref{propo2} of Appendix C.4).
 %The hyperparameter $\epsilon$ is a small positive number to serve as the threshold of \(R_q\). Since \(R_q\) is calculated by normalizing the feature volume $\rm{Vol}_q$ with $\rm{Vol}_q^{*}$, both of which depend on $q$ (i.e., $N_q^{mb}$), its threshold $\epsilon$ largely mitigates the sensitivity of $q$ to various domains and can be empirically chosen as a constant value.}

Putting $\mathcal{L}_1$ and $\mathcal{L}_2$ together, our method is finally reformulated as
\begin{equation}
 \begin{split} 
     %\max_{q} \ N_q^{mb}, s.t. \ q\in Q, \ |R_q - 1| < \epsilon, \ \left \langle q, N_q^{mb} \right \rangle = mb.
    & \max_{q} \ g(q) \\
    & s.t., \ q\in Q, \ |R_q - 1| < \epsilon,  
    % \ \left \langle q, N_q^{mb} \right \rangle = mb,
 \end{split}
 \label{Objective}
\end{equation}
where $\epsilon$ is a small positive number to serve as the threshold of \(R_q\). 
$g(\cdot): \mathbb{R} \rightarrow \mathbb{N}$ represents the function that outputs the maximum number (i.e., \({N}_{q}^{mb}\)) such that the compressed version of \({D}^{mb *}_q\) to a quality \(q\) can be stored in the memory buffer.
Note that we relax the original restriction of minimizing \(|R_q - 1|\) by enforcing \(|R_q - 1| < \epsilon\), since maximizing \(N_q^{mb}\) for $\mathcal{L}_1$ and maximizing \(q\) for $\mathcal{L}_2$ cannot be achieved simultaneously. 
Of note, \(R_q\) is calculated by normalizing the feature volume $\rm{Vol}_q^c$ with $\rm{Vol}_q^{*}$, both of which depend on $q$ (i.e., $N_q^{mb}$). Therefore, $\epsilon$ can largely mitigate the sensitivity of $q$ to various domains and can be empirically set as a constant value (see Sec. 4.3). 

Since the function $g(\cdot)$ in Eq.~(\ref{Objective}) is highly non-smooth, gradient-based methods are not applicable. Indeed, we solve it by selecting the best candidate in a finite-size set $Q$. Generally, the candidate values in $Q$ can be equidistantly selected from the range of $q$, such as \([1, 100]\) for JPEG. More candidate values can determine a proper $q$ more accurately, but the complexity will grow linearly. We found that selecting 5 candidate values is a good choice in our experiments. Once we solve Eq.~(\ref{Objective}), a good trade-off is achieved by reducing \(q\) as much as possible to obtain a larger \(N_q^{mb}\), while keeping the feature volume $\rm{Vol}_q^c$ similar to $\rm{Vol}_q^{*}$. This is consistent with our empirical analysis in Fig. \ref{Feature_Manifold_Compressed}.

\subsection{Validate Our Method with Grid Search Results}

%\begin{figure}[th]
\begin{wrapfigure}{r}{0.60\textwidth}
    \vspace{-.2cm}
	\centering
	\includegraphics[width=0.60\columnwidth]{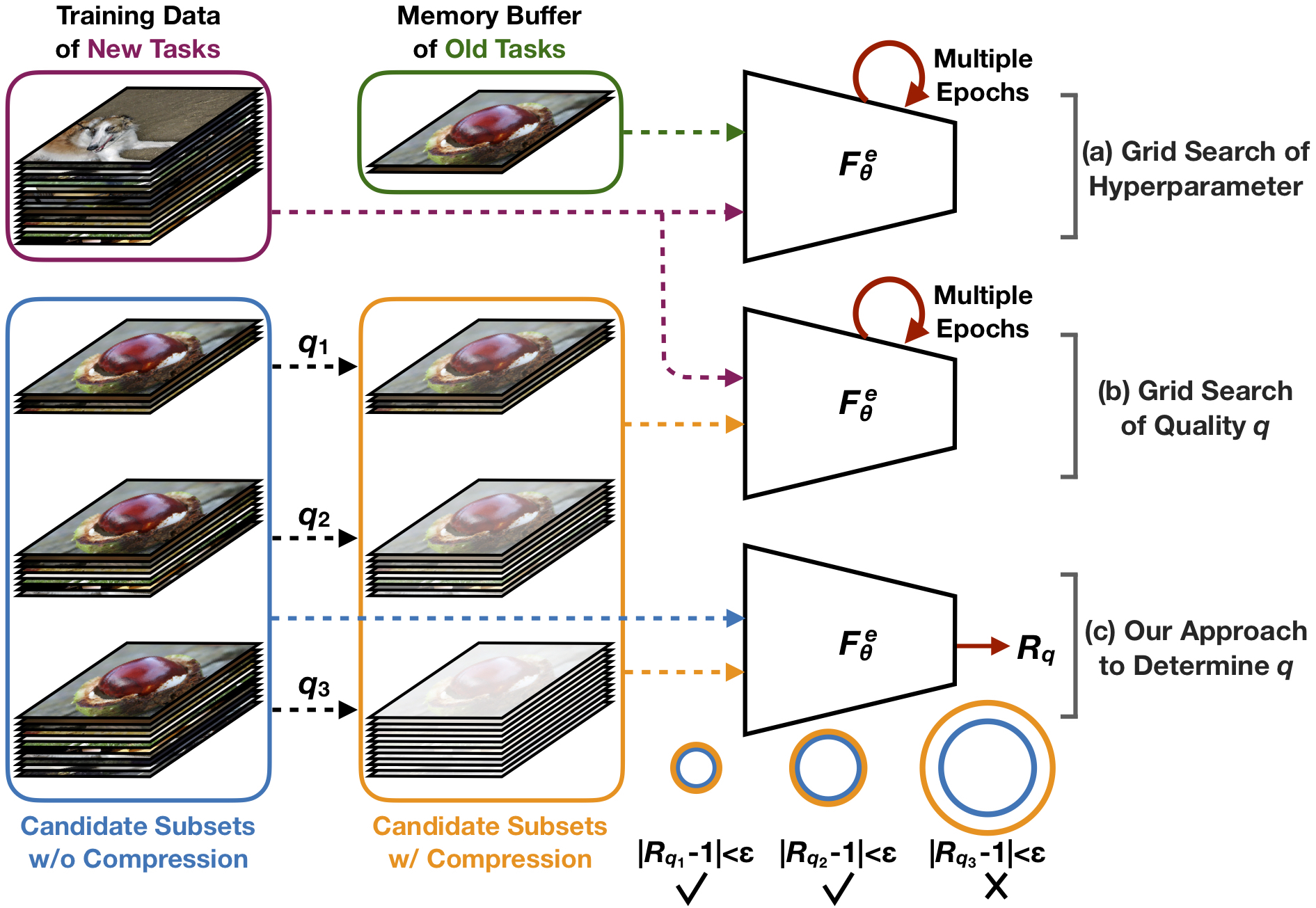} 
    \vspace{-.6cm}
	%\caption{Determine an appropriate quality efficiently.}
    \caption{Properly select a quality without repetitive training.}
	\label{fig:Determine_Compression_Rate}
    \vspace{-.4cm}
\end{wrapfigure}
%\end{figure}

In essence, the grid search described in Sec.~4.1 can be seen as a naive approach to determine the compression quality, which is similar to selecting other hyperparameters for continual learning (Fig.~\ref{fig:Determine_Compression_Rate}, a). 
This strategy is to learn a task sequence or sub-sequence using different qualities and choose the best one (Fig.~\ref{fig:Determine_Compression_Rate}, b), which leads to huge extra computation and will be less applicable if the old data cannot be revisited, or the future data cannot be accessed immediately.
%conflicts with the restriction of continual learning.
%This strategy is achieved by learning each compressed subset \(D_q^{mb}\) together with the training dataset of incoming tasks for multiple epochs (Fig.~\ref{fig:Determine_Compression_Rate}, b), resulting in huge computation cost. 
%In contrast, our method described in Sec. 4.2 can efficiently determine \(q \) without extra training, which only needs to calculate the feature volumes of each compressed subset \(D_{q}^{mb}\) and original subset \(D_{q}^{mb*}\) (Fig.~\ref{fig:Determine_Compression_Rate}, c).
In contrast, our method described in Sec. 4.2 only needs to calculate the feature volumes of each compressed subset \(D_{q}^{mb}\) and original subset \(D_{q}^{mb*}\) (Fig.~\ref{fig:Determine_Compression_Rate}, c), without repetitive training.

\begin{figure}[ht]
%\begin{wrapfigure}{r}{0.40\textwidth}
    \vspace{-.5cm}
	\centering
     \includegraphics[width=0.85\columnwidth]{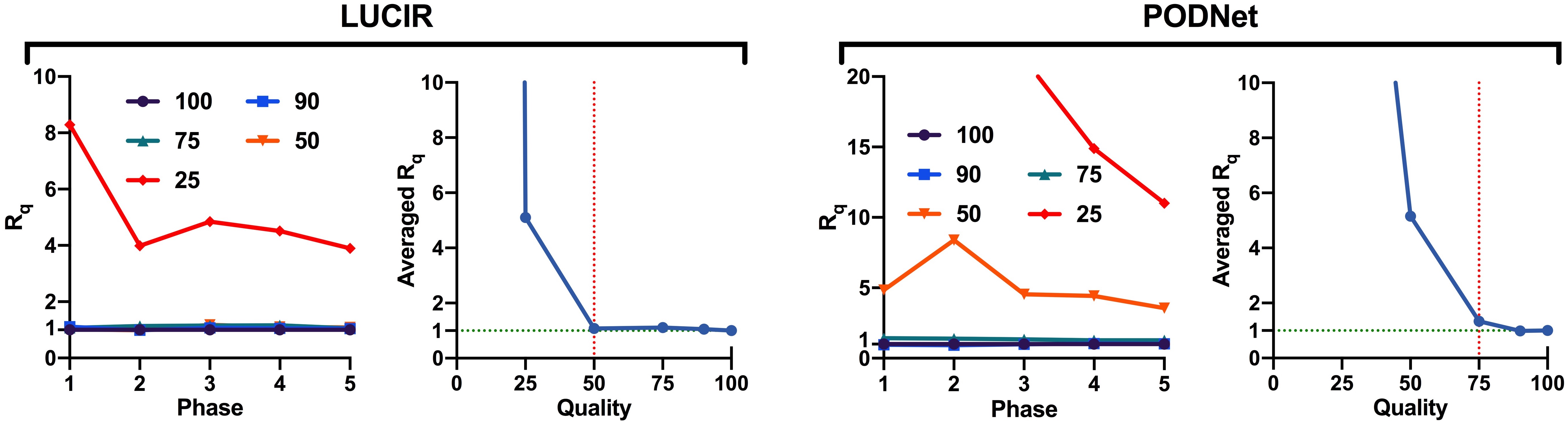}
    \vspace{-0.3cm}
    \caption{For 5-phase ImageNet-sub, we present \(R_q\) in each incremental phase with various compression qualities \(q\), and the averaged \(R_q\) of all incremental phases. } 
    \label{5phase_ImageNet_Rq}
    \vspace{-0.5cm}
%\end{wrapfigure}
\end{figure}

Now we validate the quality determined by our method with the grid search results, where LUCIR and PODNet achieve the best performance at the JPEG quality of 50 and 75 on ImageNet-sub, respectively (see Fig. \ref{Accuracy_Tradeoff}). We present \(R_q\) in each incremental phase and the averaged \(R_q\) of all incremental phases for 5-phase split in Fig. \ref{5phase_ImageNet_Rq} and for 10- and 25-phase splits in Appendix Fig.\ref{Volume_by_Phase_All}. Based on the principle in Eq.~(\ref{Objective}) with \(\epsilon = 0.5\), it can be clearly seen that 50 and 75 are the qualities chosen for LUCIR and PODNet, respectively, since they are the smallest qualities that satisfy \(|R_q - 1| < \epsilon\). Therefore, the quality determined by our method is consistent with the grid search results, but the computational cost is saved by more than 100 times.
Interestingly, for each quality \(q\), whether \(|R_q - 1| < \epsilon\) is generally consistent in each incremental phase and the average of all incremental phases. We further explore the scenarios where \(R_q \) might be more dynamic in Appendix D.4.

\section{Experiment}
In this section, we first evaluate memory replay with data compression (MRDC) in class-incremental learning of large-scale images. Then, we demonstrate the advantages of our proposal in realistic semi-supervised continual learning of large-scale object detection for autonomous driving.\footnote{All experiments are averaged by more than three runs with different random seeds.} 
%Our code is included in Supplementary Materials.
%\footnote{All results are averaged by more than three runs with different random seeds, splits of incremental phases and task orders. We include our code in Supplementary Materials.}

\subsection{Class-Incremental Learning}
\textbf{Benchmark}: We consider three benchmark datasets of large-scale images for continual learning. CUB-200-2011 \citep{wah2011caltech} is a large-scale dataset including 200-class 11,788 colored images of birds with default format of JPG, split as 30 images per class for training while the rest for testing. ImageNet-full \citep{russakovsky2015imagenet} includes 1000-class large-scale natural images with default format of JPEG.
%, with hundreds to 1300 training samples and 50 testing samples per class. %The default format of ImageNet is JPEG, while CUB-200-2011 is JPG. 
ImageNet-sub \citep{hou2019learning} is a subset derived from ImageNet-full, consisting of randomly selected 100 classes of images. Following  \cite{hou2019learning}, we randomly resize, crop and normalize the images to the size of \(224 \times 224\), and randomly split a half of the classes as the initial phase while split the rest into 5, 10 and 25 incremental phases. We report the averaged incremental accuracy with single-head evaluation \citep{chaudhry2018riemannian} in the main text, and further present the averaged forgetting in Appendix E.3.
%(detailed in Appendix B.1). We further present the metric of averaged forgetting in Appendix E.3. 

\textbf{Implementation}: 
%The implementation is the same as representative memory replay approaches \citep{hou2019learning,douillard2020podnet}, detailed in Appendix B.1. We mainly consider a memory buffer where the storage space is equal to 20 original images per class. 
We follow the implementation of representative memory replay approaches \citep{hou2019learning,douillard2020podnet} (detailed in Appendix B.1), where we focus on constraining a certain storage space of the memory buffer rather than a certain number of images. The storage space is limited to the equivalent of 20 original images per class, if not specified.
We further discuss the effects of different storage space in Fig. \ref{Buffer_Size}, a fixed memory budget in Appendix E.4 and less compressed samples in Appendix E.5. For data compression, we apply a naive but commonly-used JPEG \citep{wallace1992jpeg} algorithm to compress images to a controllable quality in the range of \([1, 100]\).

\renewcommand\arraystretch{1.5}
\begin{table*}[t]
	\centering
    \vspace{-.4cm}
    \caption{Averaged incremental accuracy (\(\%\)) of classification tasks. 
    %For CUB-200-2011, all baselines are reproduced from their officially-released code (if applicable), while for ImageNet-sub/-full we present the reported performance of all baselines and the reproduced results of AANets and DDE, so as to make the comparison as fair as possible. 
    The reproduced results are presented with \(\pm\) standard deviation, while others are reported results. The reproduced results might slightly vary from the reported results due to different random seeds. \(^1\)With class-balance finetuning. \(^2\)PODNet reproduced by \cite{hu2021distilling} underperforms that in \cite{douillard2020podnet}.} %The memory buffer size is equal to 20 original images per class. 
    %\vspace{-.1cm}
	\smallskip
	\resizebox{1\textwidth}{!}{ 
	\begin{tabular}{lcccccccc}
		\specialrule{0.01em}{1.2pt}{1.5pt}
		 \multicolumn{1}{c}{} & \multicolumn{3}{c}{CUB-200-2011} & \multicolumn{3}{c}{ImageNet-sub} & \multicolumn{2}{c}{ImageNet-full} \\
    \cmidrule(lllr){2-4}  \cmidrule(lllr){5-7} \cmidrule(lllr){8-9}
       Method & 5-phase & 10-phase & 25-phase & 5-phase & 10-phase & 25-phase & 5-phase & 10-phase \\
      \specialrule{0.01em}{1.2pt}{1.7pt}
      LwF \citep{li2017learning} &39.42 \tiny{$\pm 0.48$} &38.53 \tiny{$\pm 0.96$}&36.33 \tiny{$\pm 0.74$}& 53.62 & 47.64 & 44.32 & 44.35 & 38.90 \\
      iCaRL \citep{rebuffi2017icarl} &39.49 \tiny{$\pm 0.58$}&39.31 \tiny{$\pm 0.66$}&38.77 \tiny{$\pm 0.73$}& 65.44 & 59.88 & 52.97 & 51.50 & 46.89 \\
      BiC \citep{wu2019large} &45.29 \tiny{$\pm 0.88$}&45.25 \tiny{$\pm 0.70$}&45.17 \tiny{$\pm 0.27$}& 70.07 & 64.96 & 57.73 & 62.65 & 58.72 \\
      Mnemonics \citep{liu2020mnemonics}& -- & -- & -- & 72.58 & 71.37 & 69.74 & 64.54 & 63.01 \\
      TPCIL \citep{tao2020topology} &-- &-- &-- & 76.27 & 74.81 & -- & 64.89 & 62.88 \\
       \specialrule{0.01em}{1.2pt}{1.7pt}
       LUCIR \citep{hou2019learning}&44.63 \tiny{$\pm 0.32$}&45.58 \tiny{$\pm 0.28$}&45.48 \tiny{$\pm 0.66$}& 70.84 & 68.32 & 61.44 & 64.45 & 61.57 \\
      \quad w/ AANets \citep{liu2021adaptive} &-- &-- &-- & 72.55 & 69.22 & 67.60 & 64.94 & 62.39 \\
       \quad w/ DDE \citep{hu2021distilling}&-- &-- &-- & 72.34 & 70.20 & -- & 67.51\(^1\) & 65.77\(^1\)  \\
      % w/ \emph{Ours} (lossless) & 71.00 & 68.72 & 62.23 &  &  &  & & &  \\
      \rowcolor{black!15}
      \quad w/ \emph{MRDC (Ours)} &46.68 \tiny{$\pm 0.60$}&47.28 \tiny{$\pm 0.51$}&48.01 \tiny{$\pm 0.72$}& 73.56 \tiny{$\pm 0.27$}& 72.70 \tiny{$\pm 0.47$}&70.53 \tiny{$\pm 0.57$}&{67.53 \tiny{$\pm 0.08$}}\(^1\)&{65.29 \tiny{$\pm 0.10$}}\(^1\)\\
      \cdashline{1-9}[2pt/2pt]
     \quad w/ AANets (Reproduced) &46.87 \tiny{$\pm 0.66$}&47.34 \tiny{$\pm $ 0.77}&47.35 \tiny{$\pm 0.95$}&72.91 \tiny{$\pm 0.45$} & 71.93 \tiny{$\pm 0.52$}& 70.70 \tiny{$\pm 0.46$}&63.37 \tiny{$\pm 0.26$}&62.46 \tiny{$\pm 0.14$}\\
       \rowcolor{black!15}
      \quad w/ AANets + \emph{MRDC (Ours)}& \textbf{49.02} \tiny{$\pm 1.07$}&\textbf{49.84} \tiny{$\pm 0.87$}&\textbf{51.33} \tiny{$\pm 1.42$}&73.79 \tiny{$\pm 0.42$} &73.73 \tiny{$\pm 0.37$}&\textbf{73.47} \tiny{$\pm 0.35$}&64.99 \tiny{$\pm 0.13$}&63.04 \tiny{$\pm 0.11$}\\
      \cdashline{1-9}[2pt/2pt]
       \quad w/ DDE (Reproduced) & 45.86 \tiny{$\pm 0.65$}&46.48 \tiny{$\pm 0.69$}&46.46 \tiny{$\pm 0.33$}& 73.04 \tiny{$\pm 0.36$}& 70.84 \tiny{$\pm 0.59$}& 66.61 \tiny{$\pm 0.68$}& {66.95 \tiny{$\pm 0.09$}}\(^1\)& {65.21 \tiny{$\pm 0.05$}}\(^1\)\\
        \rowcolor{black!15}
      \quad w/ DDE + \emph{MRDC (Ours)} &47.16 \tiny{$\pm 0.60$}&48.33 \tiny{$\pm 0.48$}&48.37 \tiny{$\pm 0.34$}&75.12 \tiny{$\pm 0.17$}&73.39 \tiny{$\pm 0.29$}&70.83 \tiny{$\pm 0.34$}&{67.90 \tiny{$\pm 0.05$}}\(^1\)&{\textbf{66.67} \tiny{$\pm 0.23$}}\(^1\)\\
      \specialrule{0.01em}{1.2pt}{1.5pt}
       PODNet \citep{douillard2020podnet} &44.92 \tiny{$\pm 0.31$}&44.49 \tiny{$\pm 0.65$}&43.79 \tiny{$\pm 0.44$}& 75.54 & 74.33 & 68.31 & 66.95 & 64.13  \\ 
       \quad w/ AANets \citep{liu2021adaptive} &-- &-- &-- & 76.96 & 75.58 & 71.78 & 67.73 & 64.85\\
       \quad w/ DDE \citep{hu2021distilling} &-- &-- &-- & 76.71 & 75.41 & -- & 66.42\(^2\) & 64.71 \\
      % w/ \emph{Ours} (lossless) &77.35  & 74.43 &67.82 &  &  &  & & & \\
      \rowcolor{black!15}
       \quad w/ \emph{MRDC (Ours)} &46.00 \tiny{$\pm 0.28$}&46.09 \tiny{$\pm 0.37$}&45.84 \tiny{$\pm 0.43$}& \textbf{78.08} \tiny{$\pm 0.66$}& \textbf{76.02} \tiny{$\pm 0.54$}& 72.72 \tiny{$\pm 0.74$} &\textbf{68.91} \tiny{$\pm 0.16$}&66.31 \tiny{$\pm 0.26$}\\
    \specialrule{0.01em}{1.2pt}{1.7pt}
	\end{tabular}
	}
	\label{table:classification_result}
	\vspace{-.4cm}
\end{table*}
%\fi

\textbf{Baseline}: 
We evaluate representative memory replay approaches such as LwF \citep{li2017learning}, iCaRL \citep{rebuffi2017icarl}, BiC \citep{wu2019large}, LUCIR \citep{hou2019learning}, Mnemonics \citep{liu2020mnemonics}, TPCIL \citep{tao2020topology}, PODNet \citep{douillard2020podnet}, DDE \citep{hu2021distilling} and AANets \citep{liu2021adaptive}. In particular, AANets and DDE are the \emph{recent strong approaches} implemented on the backbones of LUCIR and PODNet, so we also implement ours on the two backbones. Since both AANets and DDE only release their official implementation on LUCIR, we further reproduce LUCIR w/ AANets and  LUCIR w/ DDE for fair comparison.

%In particular, since AANets \citep{liu2021adaptive} and DDE \citep{hu2021distilling} are the SOTA methods implemented on the backbones of LUCIR \citep{hou2019learning} and PODNet \citep{douillard2020podnet}, and only release their official implementation on LUCIR, we also implement our method on the backbones of LUCIR and PODNet while further reproduce LUCIR w/ AANets and  LUCIR w/ DDE for fair comparison.
%In particular, since many SOTA methods \citep{hu2021distilling, liu2021adaptive,liu2020mnemonics} are implemented on the backbones of previous works, such as LUCIR \citep{hou2019learning} and PODNet \citep{douillard2020podnet}, we also implement our method in the two backbone approaches.

\textbf{Accuracy:} 
We summarize the performance of the above baselines and memory replay with data compression (MRDC, ours) in Table \ref{table:classification_result}. %For class-incremental learning, AANets and DDE are the recent strong approaches, both of which are implemented on the backbones of LUCIR and PODNet. 
Using the same extra storage space, ours achieves comparable or better performance than AANets and DDE on the same backbone approaches, and can further boost their performance by a large margin.
The improvement from ours is due to mitigating the averaged forgetting, detailed in Appendix E.3.
For CUB-200-2011 with only a few training samples, all old data can be saved in the memory buffer with a high JPEG quality of 94. In contrast, for ImageNet-sub/-full, only a part of old training samples can be selected, compressed and saved in the memory buffer, where our method can efficiently determine the compression quality for LUCIR and PODNet (see Sec. 4.3 and Appendix D.2), and for AANets and DDE (see Appendix D.3).

%It can be clearly seen that using compressed data for memory replay can substantially boost the performance of LUCIR \cite{hou2019learning}, PODNet \cite{douillard2020podnet}, AANets \cite{liu2021adaptive} and DDE \cite{hu2021distilling}, and achieves the new SOTA performance. 

%\begin{figure}[th]
\begin{wrapfigure}{r}{0.60\textwidth}
    \centering
    %\vspace{-0.1cm}
    \includegraphics[width=0.95\linewidth]{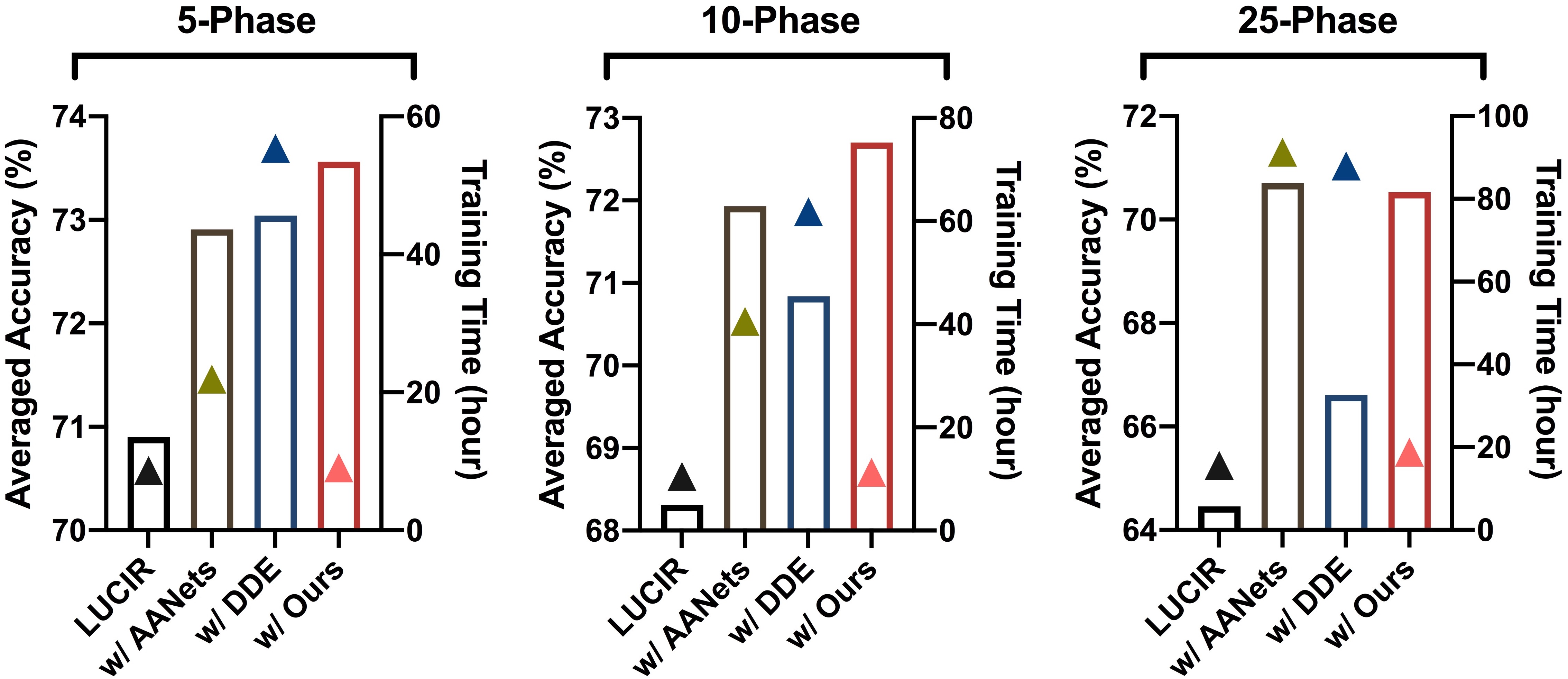}
    \vspace{-0.3cm}
    \caption{Averaged incremental accuracy (the column, left Y-axis) and computational cost (the triangle, right Y-axis) on ImageNet-sub. We run each baseline with one Tesla V100.}  
    \vspace{-0.5cm}
    \label{Time_Accuracy}
    %\vspace{-0.5cm}
\end{wrapfigure}
%\end{figure}

\textbf{Computational Cost:} 
Limiting the size of memory buffer is not only to save its storage, but also to save the extra computation of learning all old training samples again. Here we evaluate the computational cost of AANets, DDE and memory replay with data compression (MRDC, ours) on LUCIR (see Fig.~\ref{Time_Accuracy} for ImageNet-sub and Appendix Table \ref{table:time_cost} for CUB-200-2011). Both AANets and DDE require a huge amount of extra computation to improve the performance of continual learning, which is generally \emph{several times} that of the backbone approach. In contrast, ours achieves competing or more performance improvement but only slightly increases the computational cost. %(e.g., less than 20\% for ImageNet-sub and less than 40\% for CUB-200-2011).

\begin{figure}[h]
%\begin{wrapfigure}{r}{0.55\textwidth}
    \centering
    \vspace{-0.1cm}
    \includegraphics[width=1\linewidth]{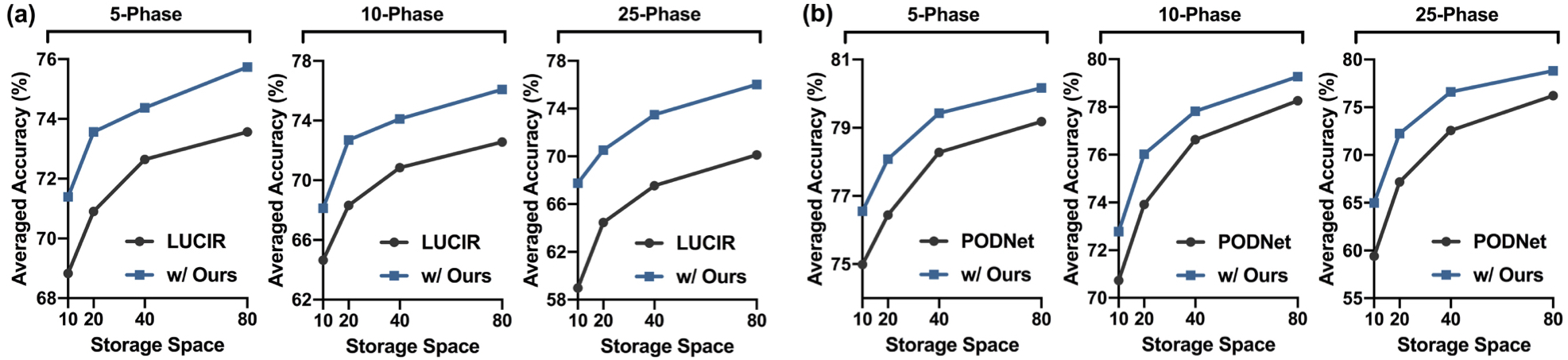}
    \vspace{-0.5cm}
    \caption{The effects of different storage space (equal to 10, 20, 40 and 80 original images per class) on ImageNet-sub. LUCIR (a) and PODNet (b) are reproduced from their officially-released codes.}  
    %\vspace{-0.1cm}
    \label{Buffer_Size}
    \vspace{-0.1cm}
%\end{wrapfigure}
\end{figure}

\textbf{Storage Space:} 
The impact of storage space is evaluated in Fig.~\ref{Buffer_Size}. % for memory replay with data compression (ours). As shown in Fig.~\ref{Buffer_Size}, 
Limiting the storage space to equivalent of 10, 20, 40 and 80 original images per class, ours can improve the performance of LUCIR and PODNet by a similar margin, where the improvement is generally more significant for more splits of incremental phases. Further, ours can achieve consistent performance improvement when using a fixed memory budget (detailed in Appendix E.4), and can largely save the storage cost without performance decrease when storing less compressed samples (detailed in Appendix E.5).
%We further explore the effects of a fixed memory budget in Appendix E.4, where the results are consistent with the dynamic one, and less compressed samples in Appendix E.5, which can largely save the storage cost to achieve a similar accuracy.
%Therefore, our method can consistently improve the efficacy of memory replay in a \emph{plug-and-play} way, without additional storage cost and additional large computation cost. 

\subsection{Large-Scale Object Detection}

%\begin{figure}[th]
\begin{wrapfigure}{r}{0.60\textwidth}
    \vspace{-.1cm}
	\centering
     \includegraphics[width=0.60\columnwidth]{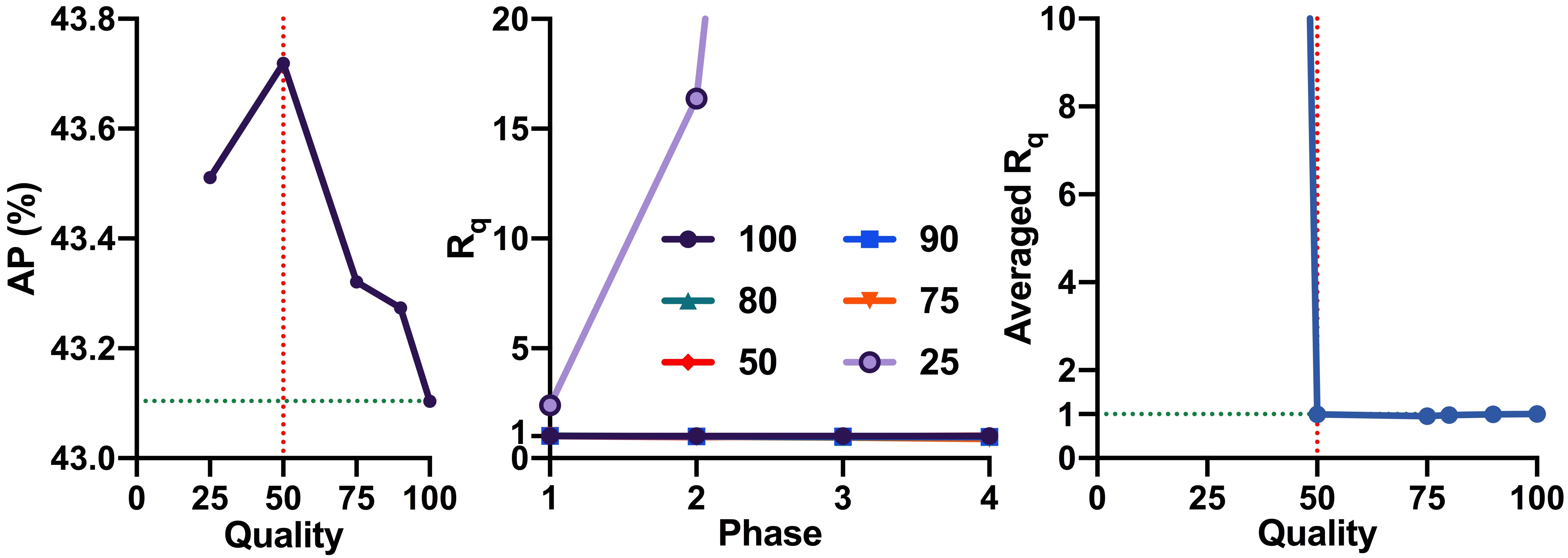}
    \vspace{-.6cm}
    \caption{We present the grid search results (left), \(R_q\) in each incremental phase (middle), and the averaged \(R_q\) of all incremental phases (right) for SSCL on SODA10M. The quality determined by our method is 50, which indeed achieves the best performance in the grid search.} 
    \vspace{-.2cm}
    \label{SODA10M_Rq}
\end{wrapfigure}
%\end{figure}

The advantages of memory replay with data compression are more significant in realistic scenarios such as autonomous driving, where the incremental data are extremely large-scale with huge storage cost. Here we evaluate continual learning on SODA10M \citep{han2021soda10m}, a large-scale object detection benchmark for autonomous driving. 
SODA10M contains 10M unlabeled images and 20K labeled images of size \(1920 \times 1080\) with default format of JPEG, which is much larger than the scale of ImageNet. The labeled images are split into 5K, 5K and 10K for training, validation and testing, respectively, annotating 6 classes of road objects for detection. %The default format of SODA10M is JPEG.

%\begin{table*}[t]
\begin{wraptable}{r}{8.4 cm}
	\centering
	\vspace{-.4cm}
	\caption{Detection results with \(\pm\) standard deviation (\%) of semi-supervised continual learning on SODA10M. FT: finetuning. MR: memory replay.}
    %\vspace{-.1cm}
	\smallskip
	\resizebox{0.60\textwidth}{!}{ 
	\begin{tabular}{ccccc}
		\specialrule{0.01em}{1.2pt}{1.5pt}
       &Method &$\rm{AP}$ &$\rm{AP}_{50}$ &$\rm{AP}_{75}$ \\
       \specialrule{0.01em}{1.2pt}{1.7pt}
       \multirow{3}*{\tabincell{c}{Pseudo \\ Labeling}}
      &FT     &40.36 {\tiny{$\pm 0.34$}}&63.83 {\tiny{$\pm 0.35$}}&43.82 {\tiny{$\pm 0.34$}}\\
      &MR   &40.75 {\tiny{$\pm 0.30$}} / +0.39&65.11 {\tiny{$\pm 0.64$}} / +1.28&43.53 {\tiny{$\pm 0.20$}} / -0.29 \\
      &Ours&\textbf{41.50 {\tiny{$\pm 0.06$}} / +1.14}&\textbf{65.36 {\tiny{$\pm 0.47$}} / +1.53}&\textbf{44.95 {\tiny{$\pm 0.22$}} / +1.13}\\
       \specialrule{0.01em}{1.2pt}{1.7pt}
      \multirow{3}*{\tabincell{c}{Unbiased \\ Teacher}}
      &FT     &42.88 {\tiny{$\pm 0.32$}}&66.70 {\tiny{$\pm 0.59$}}&45.99 {\tiny{$\pm 0.32$}}\\
      &MR   &43.10 {\tiny{$\pm 0.06$}} / +0.22 &66.88 {\tiny{$\pm 0.51$}} / +0.18 &46.62 {\tiny{$\pm 0.02$}} / +0.63 \\
      &Ours&\textbf{43.72 {\tiny{$\pm 0.25$}} / +0.84} &\textbf{67.80 {\tiny{$\pm 0.46$}} / +1.10} &\textbf{47.36 {\tiny{$\pm 0.23$}} / +1.37} \\
       \specialrule{0.01em}{1.2pt}{1.7pt}
	\end{tabular}
	}
	\label{table:SODA10M}
	\vspace{-.2cm}
\end{wraptable}
%\end{table*}

Since the autonomous driving data are typically partially-labeled, we follow the semi-supervised continual learning (SSCL) proposed by \cite{wang2021ordisco} for object detection. Specifically, we randomly split 5 incremental phases containing 1K labeled data and 20K unlabeled data per phase, and use a memory buffer to replay labeled data. The storage space is limited to the equivalent of 100 original images per phase. Following \cite{han2021soda10m}, we consider Pseudo Labeling and Unbiased Teacher \citep{liu2021unbiased} for semi-supervised object detection (the implementation is detailed in Appendix B.2). 
Using the method described in Sec. 4.2 with the same threshold (i.e., \(\epsilon=0.5\)), we can efficiently determine the compression quality for SSCL of object detection, and validate that the determined quality indeed achieves the best performance in grid search (see Fig. \ref{SODA10M_Rq}). Then, compared with memory replay of original data, our proposal can generally achieve \emph{several times} of the performance improvement on finetuning, as shown in Table \ref{table:SODA10M}.

\section{Conclusion}
In this work, we propose that using data compression with a properly selected compression quality can largely improve the efficacy of memory replay by saving more compressed data in a limited storage space. 
To efficiently determine the compression quality, we provide a novel method based on determinantal point processes (DPPs) to avoid repetitive training, and validate our method in both class-incremental learning and semi-supervised continual learning of object detection. Our work not only provides an important yet under-explored baseline, but also opens up a promising new avenue for continual learning. Further work could develop adaptive compression algorithms for incremental data to improve the compression rate, or propose new regularization methods to constrain the distribution changes caused by data compression. Meanwhile, the theoretical analysis based on DPPs can be used as a general framework to integrate optimizable variables in memory replay, such as the strategy of selecting prototypes.
In addition, our work suggests how to save a batch of training data in a limited storage space to best describe its distribution, which will motivate broader applications in the fields of data compression and data selection. 

\section*{Acknowledgements}
This work was supported by %the National Key Research and Development Program of China (Nos. 2020AAA0104304), 
 NSF of China Projects (Nos. 62061136001, 61620106010, 62076145, U19B2034, U181146), Beijing NSF Project (No. JQ19016),
 Beijing Outstanding Young Scientist Program NO. BJJWZYJH012019100020098,
 Tsinghua-Peking Center for Life Sciences, Beijing Academy of Artificial Intelligence (BAAI), Tsinghua-Huawei Joint Research Program, a grant from Tsinghua Institute for Guo Qiang, and the NVIDIA NVAIL Program with GPU/DGX Acceleration, Major Innovation \& Planning Interdisciplinary Platform for the ``Double-First Class" Initiative, Renmin University of China.

\section*{Ethic Statement}
This work presents memory replay with data compression for continual learning. It can be used to reduce the storage requirement in memory replay and thus may facilitate large-scale applications of such methods to real-world problems (e.g., autonomous driving).
As a fundamental research in machine learning, the negative consequences in the current stage are not obvious. 

\section*{Reproducibility Statement }
We ensure the reproducibility of our paper from three aspects. 
(1) Experiment: The implementation of our experiment described in Sec. 4.1, Sec. 4.3 and Sec. 5 is further detailed in Appendix B. 
(2) Code: Our code is included in supplementary materials.
(3) Theory and Method: A complete proof of the theoretical results described in Sec. 4.2 is included in Appendix C.

\clearpage

\bibliography{main}
\bibliographystyle{iclr2022_conference}

\clearpage

\appendix

\section{Trade-off between Quality and Quantity}

\begin{wrapfigure}{r}{0.60\textwidth}
%\begin{figure}[th]
    \centering
    \vspace{-0.1cm}
    \includegraphics[width=1\linewidth]{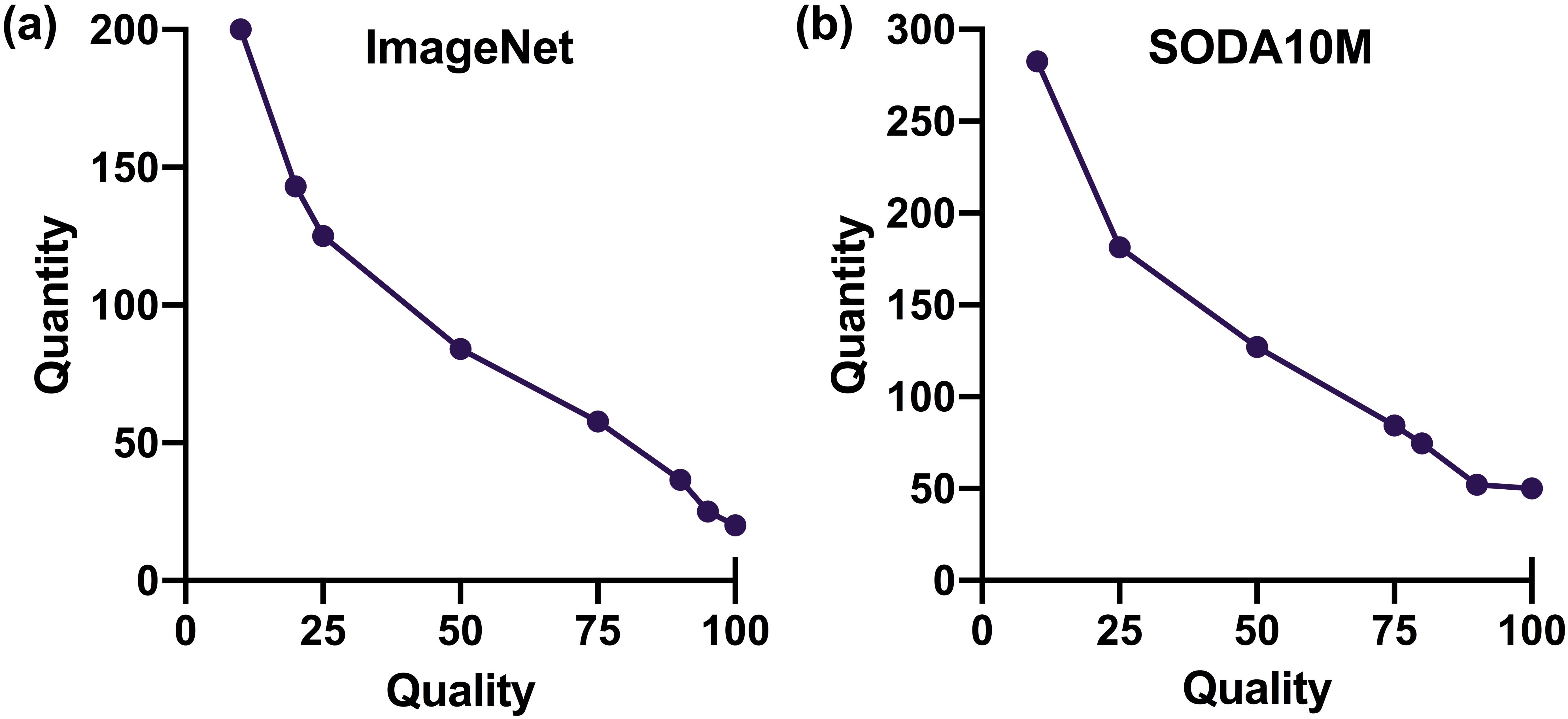}
    \vspace{-0.6cm}
    \caption{Limiting the storage space to the equivalent of (a) 20 original images per class for ImageNet and (b) 50 original images per phase for SODA10M, we plot the relation between the quality and quantity of compressed data. The quality of 100 refers to the original data without compression.}
    \label{Compression_Rate}
    \vspace{-0.2cm}
%\end{figure}
\end{wrapfigure}
\begin{figure}[th]
    \centering
    \vspace{-0.2cm}
    \includegraphics[width=0.90\linewidth]{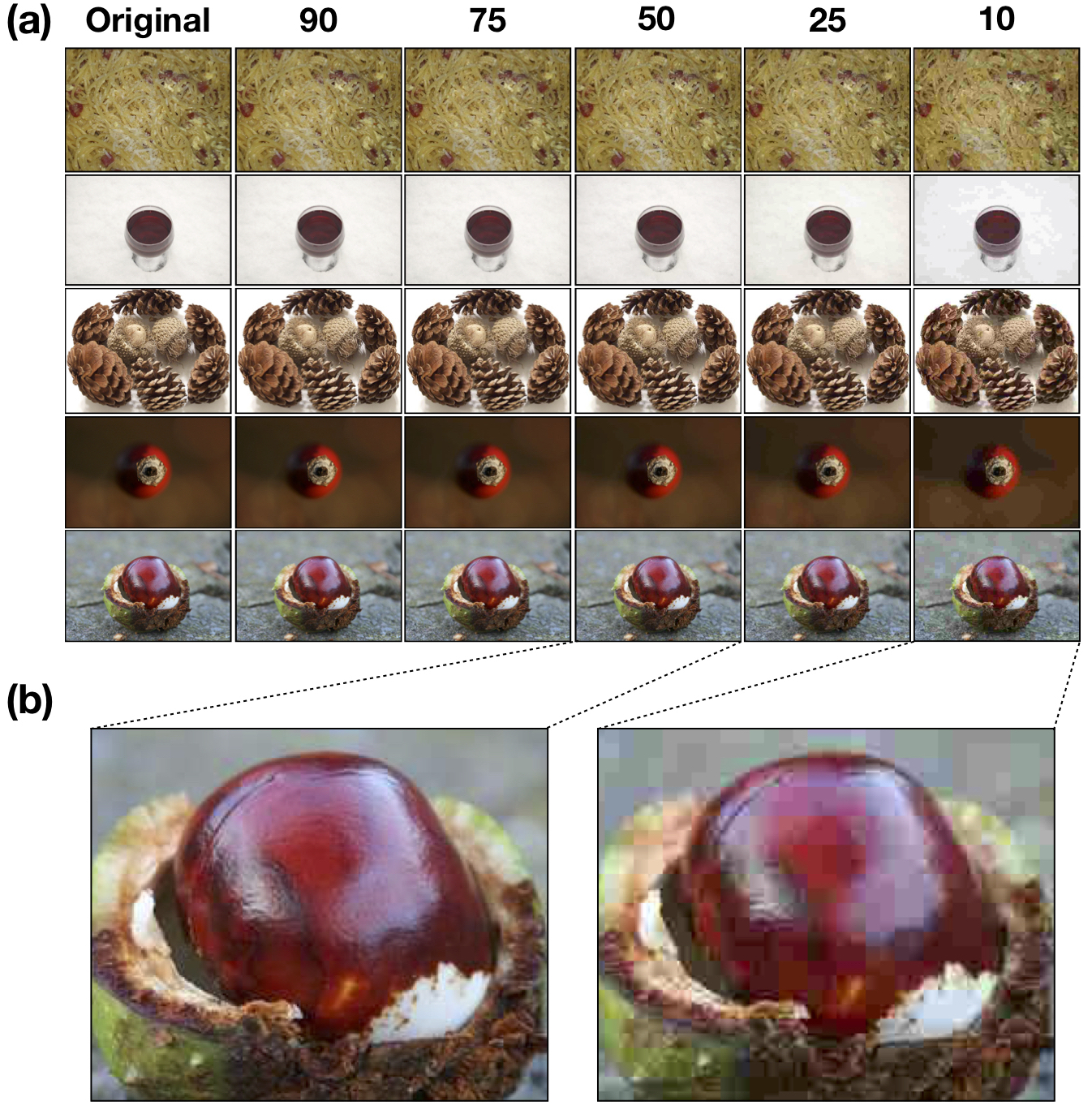}
    \vspace{-0.2cm}
    \caption{Compressed images of ImageNet. (a) Each column contains compressed images of a specific JPEG quality. (b) Two exemplar images with different qualities.}
    \label{Compressed_Images}
    \vspace{-0.2cm}
\end{figure}

Given a limited storage space, there is an intuitive trade-off between the quality \(q\) and quantity \(N_q^{mb}\) of compressed data. Here we plot the relation of \(q\) and \({N}_{q}^{mb}\) for ImageNet and SODA10M used in our paper. As shown in Fig.~\ref{Compression_Rate}, reducing the quality enables to save more compressed data in the memory buffer, and vice versa. Then, we present compressed images with various qualities in Fig.~\ref{Compressed_Images}. Although reducing the quality tends to distort the compressed data and thus hurts the performance of continual learning, it can be clearly seen that the semantic information of such compressed data is hardly affected. Therefore, a potential follow-up work is to regularize the differences between compressed data and original data, so as to further improve the performance of memory replay.

\section{Implementation Detail}
\subsection{Class-Incremental Learning}
Following the implementation of \cite{hou2019learning,douillard2020podnet,tao2020topology,liu2021adaptive,hu2021distilling}, we train a ResNet-18 architecture for 90 epochs, with minibatch size of 128 and weight decay of \(1 \times 10^{-4}\). We use a SGD optimizer with initial learning rate of 0.1, momentum of 0.9 and cosine annealing scheduling. We run each baseline with one Tesla V100, and measure the computational cost. 
We evaluate the performance of class-incremental learning with the averaged incremental accuracy (AIC) \citep{hou2019learning}. After learning each phase, we can calculate the test accuracy on all classes learned so far in the previous $t$ phases as $\rm{A}_t$. Then we can calculate \(\rm{AIC}_t = \frac{1}{t} \sum_{i=1}^t \rm{A}_i\). %All experiments are averaged by more than three runs with different random seeds and task orders.

To evaluate the effects of class similarity on the determined quality of data compression, we select similar or dissimilar superclasses from ImageNet, which includes 1000 classes ranked by their semantic similarity. Specifically, we select 10 adjacent classes as a superclass, and construct 10 adjacent (similar) or remote (dissimilar) superclasses. The index of classes selected in 10 similar superclasses is [1, 2, 3, ..., 100], while in 10 dissimilar superclasses is [1, 2, ..., 10, 101, 102, ..., 110, ..., 910]. 

\subsection{Large-Scale Object Detection}
For memory replay, we follow \cite{wang2021ordisco} to randomly select the old labeled data and save them in the memory buffer, which indeed achieves competitive performance as analyzed by \cite{chaudhry2018riemannian}. 
We follow the implementation of \cite{han2021soda10m} for semi-supervised object detection. Specifically, we use Faster-RCNN \citep{ren2015faster} with FPN \citep{lin2017feature} and ResNet-50 backbone as the object detection network for Pseudo Labeling \citep{han2021soda10m} and Unbiased Teacher \citep{liu2021unbiased}. For each incremental phase, we train the network for 10K iterations using the SGD optimizer with initial learning rate of 0.01, momentum of 0.9, and constant learning rate scheduler. The batch size of supervised and unsupervised data are both 16 images. For Pseudo Labeling, we first train a supervised model on the labeled set  in the first 2K iterations. Then we evaluate the supervised model on the unlabeled set. A bounding box with a predicted score larger than 0.7 in the evaluation results is selected as a pseudo label, which would be used to train the supervised model in the later 8K iterations in each phase. For Unbiased Teacher, we follow the default setting and change the input size to comply with SODA10M. We run each baseline with 8 Tesla V100. The performance is evaluated by the metrics ($\rm{AP}$, $\rm{AP}_{50}$, $\rm{AP}_{75}$) used in the prior works \citep{liu2021unbiased, han2021soda10m}.

\section{Theoretical Analysis}

\subsection{Determinantal Point Processes (DPPs) Preliminaries\label{pre}}

Arising in quantum physics and random matrix theory, determinantal point processes (DPPs) are elegant probabilistic models of global, negative correlations, and offer efficient algorithms for sampling, marginalization, conditioning, and other inference tasks.
% We begin with an overview of the aspects of DPPs most relevant to the machine learning
% community, emphasizing intuitions, algorithms, and computation properties.
% For instance, $\mathcal{V}$ could be a discrete and finite points processes, with $\mathcal{P}(S \subseteq Y)$ characterizes the likelihood of selecting the subset $S$.

A point process $\mathcal{P}$ on a ground set $\mathcal{V}$ is a probability measure on the power set $2^N$, where $N=|\mathcal{V}|$ is the size of the ground set.
Let $B$ be a $D \times N$ matrix with $D \geq N$. In practice, the columns of $B$ are vectors representing items in the set $\mathcal{V}$.
Denote the columns of $B$ by $B_i$ for $i=1,2,\cdots,N$. 
A sample from $\mathcal{P}$ might be the empty set, the entirety of $B$, or anything in between. $\mathcal{P}$ is called a determinantal point process if, given a random subset $Y$ drawn according to $\mathcal{P}$ ($2^N$ possible instantiations for $Y$), we have for every $S \subseteq B$,
\begin{equation}\label{dpp-MarginalPro}
    \mathcal{P}(Y=S) \propto \det(L_S),
\end{equation} 
where $S=\{B_{i_{1}},B_{i_{2}},\cdots, B_{i_{|S|}}\}$, $\mathcal{P}(Y=S)$ characterizes the likelihood of selecting the subset $S$ from $B$.
For some symmetric similarity kernel $L \in \mathbb{R} ^{N \times N}$ (e.g., $L = B^{\rm{T}}B$), where $L_S$ is the similarity kernel of subset $S$ (e.g., $L = S^{\rm{T}}S$). 
That is, $L_S$ is the submatrix sampled from $L$ using indices from $S$.
%$L$ must be real, positive semidefinite matrix $L\preceq I$ (i.e., all the eigenvalues of $L$ are between 0 and 1; 
Since $\mathcal{P}$ is a probability measure, all principal minors $\det(L_S)$ of $L$ must be non-negative, and thus $L$ itself must be positive semidefinite. These requirements turn out to be sufficient: any $L$, $0 \preceq L \preceq I$, defines a DPP.

% A point process $\mathcal{P}$ on a ground set $\mathcal{V}$ is a probability measure on the power set $2^N$, where $N=|\mathcal{V}|$ is the size of the ground set.
% A sample from $\mathcal{P}$ might be the empty set, the entirety of $\mathcal{V}$, or anything in between. $\mathcal{P}$ is called a determinantal point process if, given a random subset $Y$ drawn according to $\mathcal{P}$, we have for every $S \subseteq \mathcal{V}$,
% \begin{equation}
%     \mathcal{P}(S \subseteq Y) \propto \det(L_S),
% \end{equation}
% where $L_S \equiv [L_{ij}]_{i,j\in S}$, $\mathcal{P}(S \subseteq Y)$ characterizes the likelihood of selecting the subset $S$.
% For some symmetric similarity kernel $L \in \mathbb{R} ^{N \times N}$, where $L_S$ is the similarity kernel of subset $S$. %$L$ must be real, positive semidefinite matrix $L\preceq I$ (i.e., all the eigenvalues of $L$ are between 0 and 1; 
% Since $\mathcal{P}$ is a probability measure, all principal minors $\det(L_S)$ of $L$ must be non-negative, and thus $L$ itself must be positive semidefinite. These requirements turn out to be sufficient: any $L$, $0 \preceq L \preceq I$, defines a DPP.

$L$ is often referred to as the marginal kernel since it contains all the information needed to compute the probability of any subset $S$ being included in $Y$.
Hence, the marginal probability of including one element $B_i$ is $\mathcal{P}(B_i \in Y) = L_{ii}$, and two elements $B_i$ and $B_j$ is $ \mathcal{P}(B_i, B_j \in Y) = L_{ii} L_{jj}-L_{ij}^2 = \mathcal{P}(B_i \in Y)\mathcal{P}(B_j \in Y)-L_{ij}^2$.
A large value of $L_{ij}$ reduces the likelihood of both elements to appear together in a diverse subset.
This demonstrates why DPPs are ``diversifying". 
Below is several important conclusions about DPPs related to our work~\citep{kulesza2012determinantal}.

\paragraph{Geometric Interpretation.}
By focusing on L-ensembles in DPPs, determinants have an intuitive geometric interpretation.
If $L = B^{\top}B$,
% Let $B$ be a $D \times N$ matrix such that $L = B^{\rm{T}}B$. Such a $B$ can always be found for $D\leq N$ when $L$ is positive semi-definite. In practice, the columns of $B$ are vectors representing items in the set $\mathcal{V}$.
% Denote the columns of $B$ by $B_i$ for $i=1,2,\cdots,N$. 
then
\begin{equation} 
    % \mathcal{P}(S \subseteq Y) \propto \det(L_S) ={\rm{Vol}}^{2}(\{B_i\}_{i\in S}),
    \mathcal{P}(Y=S) \propto \det(L_S) =({\rm{Vol}}(S))^{2},
\end{equation}
% $\mathcal{P}(S \subseteq Y) \propto \det(L_S) ={\rm{Vol}}^{2}(\{B_i\}_{i\in S})$,
% Then, $\mathcal{P}_L(Y)\propto \det(L_Y)=\rm{Vol}^{2}(\{B_i\}_{i\in Y})$, 
where the first two terms specify the marginal probabilities for every possible subset $S$, and the last term is the squared $|S|$-dimensional volume of the parallelepiped spanned by the items in $S$.

\paragraph{Quality vs. Diversity.}
While the entries of the DPP kernel $L$ are not totally opaque in that they can be seen as measures of similarity—
reflecting our primary qualitative characterization of DPPs as diversifying processes. In most
practical situations, we want diversity to be balanced against some underlying preferences
for different items in $B$. \cite{kulesza2012determinantal} proposed a decomposition of the DPP that more
directly illustrates the tension between diversity and a per-item measure of quality.
Specifically, $L = B^{\top}B$ can be taken one step further, by writing each column $B_i$ as the product of a quality term $q_i \geq 0$ and  vector of normalized diversity features $f_i \in \mathbb{R}^{D}$, $\left \| f_i \right \|=1$. 
Then, $L_{ij}=q_i f_i^{\rm T} f_j q_j$, where
$f_i^{\top} f_j \in [-1, 1]$ is a signed measure of similarity between $B_i$ and $B_j$. Let
$\phi_{ij} = f_i^{\top} f_j = \frac{L_{ij}}{\sqrt{L_{ii}L_{jj}}}$, and $\phi=\{\phi_{ij}\}_{i,j=1}^{N}$.
Then, depending on Eq.~(\ref{dpp-MarginalPro}), we have 
\begin{equation}
    \mathcal{P}(Y=S) \propto \left (  \prod_{B_{i_{j}} \in S} q_{i_{j}}^2 \right )\det (\phi_S), %\phi(S)^{\rm T}\phi(S)
\end{equation}
where $q_{i_{j}}$ can be seen as a quality score of an item $B_{i_{j}}$ in $B$, and 
$\phi_S$ is the submatrix sampled from $\phi$ using indices from $S$.
%  $f_j^{\rm T} f_k \in [-1, 1]$ is a signed measure of similarity between $B_{i_{j}}$ and $B_{i_{k}}$, and
% $\phi_{jk} = f_j^{\top} f_k = \frac{L_{ij}}{\sqrt{L_{ii}L_{jj}}}$.

\paragraph{Alternative Likelihood Formulas.}
In an L-ensemble DPP, the likelihood of a particular set $S \subseteq B $ is given by
\begin{equation}
     \mathcal{P}_L(S) = \frac{\det(L_S)}{\sum_{S'} \det{(L_{S'})}} = \frac{\det(L_S)}{\det(L+I)},
\end{equation}
where $S'$ is one of all $2^N$ possible subsets in $B$, and $L_{S'}$ is the submatrix sampled from $L$ using indices from $S'$.

This expression has some nice intuitive properties in terms of volumes, and, ignoring the normalization in the denominator, takes a simple and concise form. However, as a ratio of determinants on matrices of differing dimension, it may not always be analytically convenient.
Minors can be difficult to reason about directly, and ratios complicate calculations like derivatives. Such a high computational cost is just one key challenge in DPPs. 

\paragraph{Conditional DPPs.}
A conditional DPP $\mathcal{P}(Y = S|B)$ is a conditional probabilistic model which assigns a probability to every possible subset $S \subseteq  B$. The model takes the form of an L-ensemble:
\begin{equation}
    \mathcal{P}(Y = S|B) \propto \det(L_S(B)),
\end{equation}
where $L(B)$ is a positive semidefinite conditional $N \times N$ kernel matrix that depends on the input $B$.
% A conditional DPP $\mathcal{P}(Y = S|X)$ is a conditional probabilistic model which assigns a probability to every possible subset $S \subseteq \mathcal{V}(X)$. The model takes the form of an L-ensemble:
% \begin{equation}
%     \mathcal{P}(Y = S|X) \propto \det(L_S(X)),
% \end{equation}
% where $L(X)$ is a a positive semidefinite $|\mathcal{V}(X)| \times |\mathcal{V}(X)|$ kernel matrix that depends on the input.

\paragraph{$k$-DPPs.}
% A $k$-DPP is obtained by conditioning a standard DPP on the event that the set $Y$ has cardinality $k$, which is concerned only with the content of a random $k$-set. Formally, the $k$-DPP $\mathcal{P}_L^k$ gives probabilities
% \begin{equation}
% \mathcal{P}_L^k(Y) = \frac{\det(L_Y)}{\sum_{|Y'|=k} \det(L_{Y'})},
% \end{equation}
% where $|Y| = k$ and $L$ is a positive semidefinite kernel. 
A $k$-DPP is obtained by conditioning a standard DPP on the event that the set $Y$ has cardinality $k$, which is concerned only with the content of a random $k$-set (i.e., without the size $k$)~\citep{kulesza2012determinantal}. Formally, the $k$-DPP $\mathcal{P}_L^k$ gives the probability of a particular set $S$ as
\begin{equation}\label{k-dpp}
\mathcal{P}_L^k(Y=S) = \frac{\det(L_S)}{\sum_{|S'|=k} \det(L_{S'})},
\end{equation}
where $|S| = k$ and $L$ is a positive semidefinite kernel. Although numerous optimization algorithms have been proposed to solve DPPs problems, the high computational complexity about denominator cannot be avoided generally. On the one hand, computing denominator is a sum over $\binom N k$ terms. On the other hand, computing the determinant of each term through matrix decomposition is with $O(k^3)$ time. Just as \cite{kulesza2012determinantal} claimed, sparse storage of larger
matrices is possible for DPPs, but computing determinants remains prohibitively expensive unless the level of sparsity is extreme.

If the dot product kernel function is adopted to compute the kernel $L$ in Eq.~(\ref{k-dpp}), similar to Geometric Interpretation of standard DPPs above, the probability $\mathcal{P}_L^k(Y=S)$ is further defined with volume sampling~\citep{deshpande2010efficient}, i.e., 
\begin{equation} \label{kdpp-vol}
\mathcal{P}_L^k(Y=S) = \frac{\det(L_S)}{\sum_{|S'|=k} \det(L_{S'})} \propto \det(L_S) = (k!\cdot (\rm{Vol}(\rm{Conv({\bar{0}\cup {S}}})))^2),
\end{equation}
where $L = B^{\top}B$, \rm{Conv}($\cdot$) denotes the convex hull, and \rm{Vol}($\cdot$) is the $k$-dimensional volume of such a convex hull.

\subsection{Modeling Our Case as A Conditional $N_q^{mb}$-DPP}

%\begin{figure}[th]
\begin{wrapfigure}{r}{0.40\textwidth}
    \vspace{-.2cm}
	\centering
	\includegraphics[width=0.30\columnwidth]{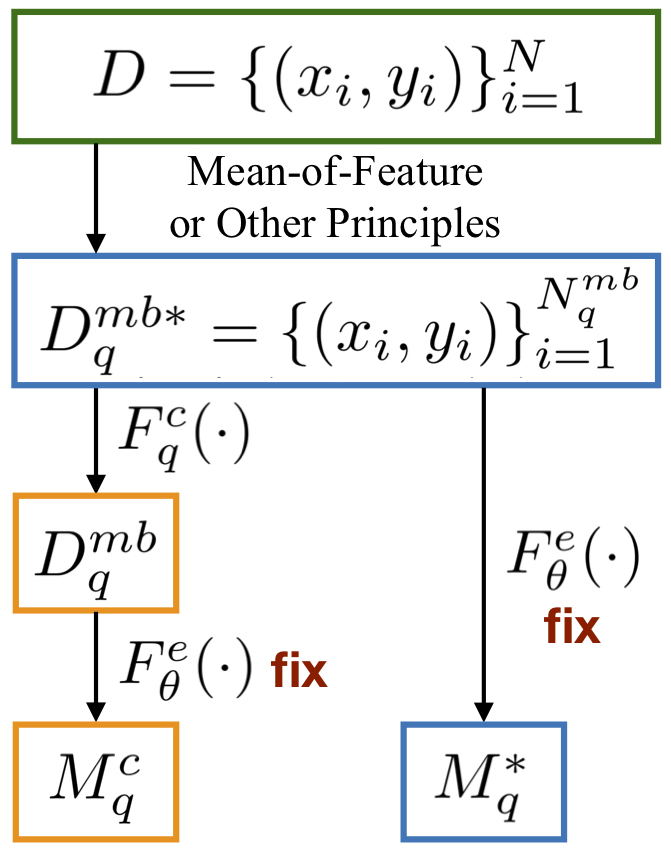} 
    \vspace{-.1cm}
	\caption{Construction of $D_q^{mb*}$, $D_q^{mb}$, $M_q^*$ and $M_q^c$. $\theta$ is fixed in this stage.} 
	\label{fig:Construct_Feature}
    \vspace{-.3cm}
\end{wrapfigure}
%\end{figure}

In this work, we aim to find the compressed subset \(D_q^{mb}\) that can best represent the training dataset \(D\) by choosing an appropriate compression quality \(q\) from its range. To achieve this goal, we introduce $\mathcal{P}_{q} (D_q^{mb}|D)$ to characterize the conditional likelihood of selecting \(D_q^{mb}\) given input \(D\) under parameter $q$. Since the network parameter $\theta$ is fixed during compression, and the feature matrix $M_q^c=F_{\theta}^e(D_q^{mb})$, we rewrite $\mathcal{P}_{q} (D_q^{mb}|D)$ as $\mathcal{P}_{q} (M_q^c|D)$ equivalently (see Fig.~\ref{fig:Construct_Feature} for the details of constructing $D_q^{mb}$ and $M_q^c$). 
Depending on the nice properties of DPPs in Sec.~\ref{pre}, 
we formulate our goal as a conditional DPP, where $D$ is the associated ground set, and $M_q^c$ is a desired subset in feature space.
% we can formulate $(D, M_q^c) \in \mathcal{X} \times 2^{|D|}$, where $\mathcal{X}$ is an input space, $\mathcal{V}(D)$ is the associated ground set for input $D$, and $M_q^c$ is a possible subset of $D$. 
Then $\mathcal{P}_{q} (M_q^c|D)$ is a distribution over all subsets of $D$ with cardinality $N_q^{mb}$, which is formally called a conditional $N_q^{mb}$-DPP, since $|{M_q^c}|=N_q^{mb}$ for any possible subset. Thus, such a DPP formulates the conditional probability $\mathcal{P}_{q} (M_q^c|D)$ as
\begin{equation}\label{Appendix:Compress-def} 
\begin{split}
  \mathcal{P}_{q} (M_q^c|D) 
  = \frac{\det(L_{M_q^c}(D; q,\theta))}{\sum_{|{M}|=N_q^{mb}}\det(L_{{M}}(D; q, \theta))},
  \end{split}
\end{equation}
where \(L(D; q, \theta) \) is a conditional DPP $|D|\times |D|$ kernel matrix that depends on the input $D$ parameterized in terms of parameters $\theta$ and $q$. $L_{M}(D; q, \theta)$ (resp., $L_{M_q^c}(D; q,\theta))$) is the submatrix sampled from \(L(D; q, \theta) \) using indices from $M$ (resp., $M_q^c$).
%The numerator defines the marginal probability of inclusion for the subset ${M_q^c}$, and the denominator serves as a normalizer to enforce the sum of $\mathcal{P}_{q} (M_q^c|D)$ for every possible ${M_q^c}$ to 1. Generally, there are many ways to obtain a positive semi-definite kernel $L$. In this work, we employ the most widely used dot product kernel function, where $L_{M_q^c}=M_q^{c \top} M_q^{c}$ and  $L_{{M}}={M}^{ \top} {M}$.

Similarly, $\mathcal{P}_{q} (M_q^*|D)$ can be formulated by a conditional $N_q^{mb}$-DPP as
\begin{equation}\label{Appendix:Uncompess-def}
\begin{split}
  \mathcal{P}_{q} (M_q^*|D)  = \frac{\det(L_{M_q^*}(D; \theta))}{\sum_{|M^*|=N_q^{mb}} \det(L_{M^*}(D; \theta))}, \\
  \end{split}
\end{equation}
where $|M_q^*|=N_q^{mb}$. Unlike \(L(D; q,\theta) \) above, the conditional DPP kernel matrix \(L(D; \theta) \) only depends on $D$ and $\theta$ without $q$, since $q$ is fixed at its maximum, i.e., without compression. $L_{M^*}(D; \theta)$ (resp., $L_{M_q^*}(D; \theta)$) is the submatrix sampled from \(L(D; \theta) \) using indices from $M^*$ (resp., $M_q^*$).

\textbf{Differences:} Of note, two important differences between standard $k$-DPPs and our $N_q^{mb}$-DPP lie in the optimization variable and objective. For this work, we need to find an optimal set cardinality (i.e., $N_q^{mb}$), while it is fixed for standard $k$-DPPs. Although both of them can finally determine a desired subset with the maximum volume, our $N_q^{mb}$-DPP can uniquely determine it by $N_q^{mb}$, while standard $k$-DPPs obtain it by maximizing Eq.~(\ref{k-dpp}).

\subsection{The First Goal}

First, we need to maximize
\begin{equation}\label{Appendix:Uncompess}
\begin{split}
  \mathcal{L}_1(q) = \mathcal{P}_{q} (M_q^*|D) 
 & = \frac{\det(L_{M_q^*}(D; \theta))}{\sum_{|M^*|=N_q^{mb}} \det(L_{M^*}(D; \theta))}\\
 &  \propto \det({M}_q^{* \top} M_q^{*}) = (\rm{Vol}_q^{*})^{2}\\
 &  = (N_q^{mb}!\cdot (\rm{Vol}(\rm{Conv}({\bar{0}}\cup M_q^*)))^2),
%   \mathcal{P}_{q} (M_q^*|D) = \frac{\det(L_{M_q^*}(D; q,\theta))}{\det(L(D; q,\theta)+I))}  = \frac{\det(L_{M_q^*}(D; q,\theta))}{\det(L(D; \theta)+I))} \propto \det(L_{M_q^*}(D; q, \theta)) = \rm{Vol}_q^{*2},
    %\mathcal{P}_{q} (M_q^*|D) &= \mathcal{P}_{q, \theta} (M_q^*|D) = \frac{\det(L_{M_q^*}(D; q,\theta))}{\det(L(D; q,\theta)+I))} \\
   % & \propto \det(L_{M_q^*}(D; q,\theta)) = \rm{Vol}_q^{*2},
  \end{split}
\end{equation}
%where $|M_q^*|=N_q^{mb}$ and \(L(D; \theta) \) is a conditional DPP $|D|\times |D|$ kernel matrix that depends on the input $D$ and parameter $\theta$. $L_{M_q^*}=M_q^{* \top} M_q^{*}$ is the inner product of $M_q^{*}$. Likewise, $L_{\widetilde{M}_q^*}=\widetilde{M}_q^{* \top} \widetilde{M}_q^*$. $\sum_{|\widetilde{M}_q^*|=N_q^{mb}} \det(L_{\widetilde{M}_q^*}(D; \theta))$ serves as a normalizer that depends on \(q\).
where $|M_q^*|=N_q^{mb}$. \(L(D; \theta) \) is a conditional DPP $|D|\times |D|$ kernel matrix that depends on the input $D$ parameterized in terms of $\theta$. 
$L_{M^*}(D; \theta)$ (resp., $L_{M_q^*}(D; \theta)$) is the submatrix sampled from \(L(D; \theta) \) using indices from $M^*$ (resp., $M_q^*$).
The numerator defines the marginal probability of inclusion for the subset ${M_q^*}$, and the denominator serves as a normalizer to enforce the sum of $\mathcal{P}_{q} (M_q^*|D)$ for every possible ${M_q^*}$ to 1. Here we employ the commonly-used dot product kernel function, where $L_{M_q^*}=M_q^{* \top} M_q^{*}$ and  $L_{{M}^*}={M}^{* \top} {M}^*$. \rm{Conv}($\cdot$) denotes the convex hull, and \rm{Vol}($\cdot$) is the $N_q^{mb}$-dimensional volume of such a convex hull. 

% where $L_{M_q^*}=M_q^{*T} M_q^{*}$ is the inner product of $M_q^{*}$ and \(L(D; \theta) \) is a conditional DPP kernel independent of \(q\) (detailed in Appendix C).
%Of note, the first equality follows because the parameter $\theta$ is fixed when we optimize $q$. 

Eq.~(\ref{Appendix:Uncompess}) provides a geometric interpretation that the conditional probability $\mathcal{P}_{q} (M_q^*|D)$ is proportional to the squared volume of \(M_q^*\) with respect to $N_q^{mb}$, denoted as $\rm{Vol}_q^{*}$. %Then we need to maximize
%  \begin{equation}
% \begin{split}
%   \mathcal{L}_1(q)& =
%   %\log \mathcal{P}_q (M_q^*|D)  = \log  \det(L_{M_q^*}(D; q,\theta)) -\log \det(L(D; q,\theta)+I).
%       \det(M_q^{*T} M_q^{*}) = k!\cdot \rm{Vol}(\rm{Conv}({\bar{0}}\cup M_q^*))^2.
% \end{split}
% \end{equation}
% In essence, optimizing $\mathcal{L}_1$ corresponds to maximizing the volume of \(M_q^*\), i.e., \(\rm{Vol}_q^*\), in terms of the quality $q$. 
For our task, $\mathcal{P}_{q} (M_q^*|D) $ monotonically increases with $N_q^{mb}$. Thus, optimizing $\mathcal{L}_1$ is converted into $\max_{q} \ N_q^{mb}$ equivalently (detailed in Proposition~\ref{propo1} as below). 
% For our task, $\mathcal{P}_{q} (M_q^*|D) $ is monotonically increasing with the increase of $N_q^{mb}$ (detailed in Lemma 1 of Appendix C). Thus, optimizing $\mathcal{L}_1$ is converted into $\max_{q} \ N_q^{mb}$ equally. 

\begin{proposition}\label{propo1}
% Given the following two optimization programs, 
% \begin{equation}\label{Uncompess}
% \begin{split}
% \min_{q} \mathcal{L}_1(q) = 
%  & = \frac{\det(L_{M_q^*}(D; q,\theta))}{\sum_{|M_q|=N_q} \det(L_{M_q}(D; \theta))}\\
%  &  \propto \det(M_q^{*T} M_q^{*})\\
%  &  = (N_q!\cdot \rm{Vol}(\rm{Conv}({\bar{0}}\cup M_q^*))^2),
%   \end{split}
% \end{equation}
For any matrix $X=({\bm{x}_1,\bm{x}_2,\cdots,\bm{x}_n})$, where $\bm{x}_i\in \mathbb R^{d}$ is the $i$-th column of $X$, $d>n$ and $X^{\top}X \neq I$, define $\mathcal{M}=\{i_1,i_2,\cdots,i_m\}$ as a subset of $[n]$ containing $m$ elements, and $X_{\mathcal{M}}$ as the submatrix sampled from $X$ using indices from $\mathcal{M}$. 
Then for 
% \begin{align*}
% P(\mathcal{M}):= \frac{\det\left(X_{\mathcal{M}}^{\top}X_{\mathcal{M}}\right)}{\sum_{|{M}'|=m} \ \det\left(X_{\mathcal{{M}'}}^{\top}X_{\mathcal{{M}'}}\right)},
% \end{align*}
\begin{align*}
P(\mathcal{M}):= \frac{\det\left(X_{\mathcal{M}}^{\top}X_{\mathcal{M}}\right)}{\sum_{|{\mathcal{M}}'|=m} \ \det\left({X}_{{\mathcal{{M}}'}}^{\top}{X}_{{\mathcal{{M}}'}}\right)},
\end{align*}
we can ensure that the following two optimization programs 
\begin{align*}
\max_{m} P(\mathcal{M})
\end{align*}
and
\begin{align*}
\max_{m} |\mathcal{M}|
\end{align*}
are equivalent. 
\end{proposition}

Below is one critical lemma for the proof of Proposition~\ref{propo1}.

\begin{lemma}\label{lemma1}
For any matrix $X=({\bm{x}_1,\bm{x}_2,\cdots,\bm{x}_n})$, where $\bm{x}_i\in \mathbb R^{d} $ is the $i$-th column of $X$ and $d>n$, define $\mathcal{M}=\{i_1,i_2,\cdots,i_m\}$ as a subset of $[n]$ containing $m$ elements, and $X_{\mathcal{M}}$ as the submatrix sampled from $X$ using indices from $\mathcal{M}$. 
Then for 
\begin{align*}
P(\mathcal{M}):= \frac{\det\left(X_{\mathcal{M}}^{\top}X_{\mathcal{M}}\right)}{\sum_{|{\mathcal{M}}'|=m} \ \det\left(X_{\mathcal{{M}'}}^{\top}X_{\mathcal{{M}'}}\right)},
\end{align*}
we can ensure that
\begin{align*}
[n] = \text{argmax}_{\mathcal{M}}P(\mathcal{M}).
\end{align*}
This means when $|\mathcal{M}| = n$, the probability $P(\mathcal{M})$ takes the maximum value.
\end{lemma}

\noindent\emph{Proof.}
From the definition, we have 
\begin{align*}
0 &  \leq  P(\mathcal{M}) = \frac{\det\left(X_{\mathcal{M}}^{\top}X_{\mathcal{M}}\right)}{\sum_{|{\mathcal {M}}'|=m} \ \det\left(X_{\mathcal{{M}'}}^{\top}X_{\mathcal{{M}'}}\right)} \\
&=\frac{\det\left(X_{\mathcal{M}}^{\top}X_{\mathcal{M}}\right)}{\det\left(X_{\mathcal{M}}^{\top}X_{\mathcal{M}}\right)+{\sum_{\{|{\mathcal {M}}'|=m\} \cap  \{\mathcal{M}'\neq \mathcal{M}\}}\ \det\left(X_{\mathcal{{M}'}}^{\top}X_{\mathcal{{M}'}}\right)}}
\leq 1.
\end{align*}
% If $|{M}'|=m<n$, we have $\binom n m$ possible options for subset $X_{\mathcal{{M}'}}$ , i.e., $\{X_{\mathcal{{M}'}_1},X_{\mathcal{{M}'}_2},\cdots,X_{\mathcal{{M}'}_{\binom n m}} \}$, where we can always obtain a sort of $0 < \det\left(X_{\mathcal{{M}'}_i}^{\top}X_{\mathcal{{M}'}_i}\right) <  \det\left(X_{\mathcal{{M}'}_j}^{\top}X_{\mathcal{{M}'}_j}\right) < \det\left(X_{\mathcal{{M}'}_k}^{\top}X_{\mathcal{{M}'}_k}\right) <\cdots$.
% That is, there always exist at least two submatrices satisfying $\det\left(X_{\mathcal{{M}'}_j}^{\top}X_{\mathcal{{M}'}_j}\right)$, $\det\left(X_{\mathcal{{M}'}_k}^{\top}X_{\mathcal{{M}'}_k}\right)>0$.
If $|\mathcal{{M}'}|=m<n$, we have $\binom n m > 1 $ possible options for subset $X_{\mathcal{{M}'}}$, where $\det\left(X_{\mathcal{{M}'}}^{\top}X_{\mathcal{{M}'}}\right) \geq 0$.
Only if $|\mathcal{M}| = n$, then $\binom n m =1$ and the probability $P(\mathcal{M})$ will definitely take the maximum value, i.e.  $P(\mathcal{M})=1$. In addition, $X^{\top}X$ has full rank due to $d>n$, and then $\det(X^{\top}X) > 0$ always holds. That is, $P(\mathcal{M})=1$ can be guaranteed only when selecting all $n$ items together.
This means $[n]= \text{argmax}_{\mathcal{M}}P(\mathcal{M})$.
It completes the proof.

\noindent \textbf{Proof of Proposition~\ref{propo1}} From the definition in ~\cite{deshpande2010efficient}, we have
\begin{equation}
\begin{split}
P(\mathcal{M})&
=\frac{\det\left(X_{\mathcal{M}}^{\top}X_{\mathcal{M}}\right)}{\sum_{|{\mathcal{M}}'|=m} \ \det\left({X}_{{\mathcal{{M}}'}}^{\top}{X}_{{\mathcal{{M}}'}}\right)}\\
% = \frac{\det\left(X_{\mathcal{M}}^{\top}X_{\mathcal{M}}\right)}{\sum_{|\tilde{M}|=m} \ \det\left(X_{\tilde{\mathcal{{M}}}}^{\top}X_{\tilde{\mathcal{{M}}}}\right)} \\
 &  \propto \det\left(X_{\mathcal{M}}^{\top}X_{\mathcal{M}}\right) \\
 &  = (m!\cdot (\rm{Vol}(\rm{Conv}({\bar{0}}\cup {X_{\mathcal{M}}})))^2),
\end{split}
\end{equation}
where \rm{Conv}($\cdot$) denotes the convex hull, and \rm{Vol}($\cdot$) is the $m$-dimensional volume of such a convex hull. 
Here, from the monotonicity side, it can be easily verified that for $X_{\mathcal{M}} \cup \bm{x}_{i_{m+1}}$, where $\bm{x}_{i_{m+1}} \in \complement_X X_{\mathcal{M}}$, 
we have 
\begin{align*}
    (m!\cdot (\rm{Vol}(\rm{Conv}({\bar{0}}\cup {X_{\mathcal{M}}}^*)))^2) < 
    ((m+1)!\cdot (\rm{Vol}(\rm{Conv}({\bar{0}}\cup X_{\mathcal{M}} \cup \bm{x}_{i_{m+1}})))^2),
\end{align*}
which means $P(\mathcal{M})$ is monotonically increasing with the increase of $|\mathcal{M}|$. 

From the extremum side, by using Lemma~\ref{lemma1}, we can ensure that the following two optimization programs 
\begin{align*}
\max_{m} P(\mathcal{M})
\end{align*}
and
\begin{align*}
\max_{m} |\mathcal{M}|
\end{align*}
have the same optimal value.
Thus, maximizing $P(\mathcal{M})$ with respect to $|\mathcal{M}|$ can be converted into $\max_{m} \ |\mathcal{M}|$ equivalently.

\subsection{The Second Goal}

Second, we need to minimize 
\begin{equation}
\begin{split}
    \mathcal{L}_2(q)& =
     \left |  \frac{ \mathcal{P}_q (M_q^{c}|D) }{ \mathcal{P}_q (M_q^{*}|D)} -1 \right | 
    =  \left |  \frac{ \det(L_{M_q^c}(D; q,\theta)) }{\det(L_{M_q^{*}}(D; \theta))}Z_q -1 \right | \\
    % & = \left | \log \left (\sqrt{\frac{\det(M_q^{\rm T} M_q)}{\det(M_q^{*T} M_q^{*})}}  \right ) -1 \right | \\
    & =  \left | \frac{\det(M_q^{c \top} M_q^c)}{\det(M_q^{* \top} M_q^{*})}Z_q  -1 \right |
    % & = \log \left | \sqrt{\frac{\det(M_q^{\rm T} M_q)}{\det(M_q^{*T} M_q^{*})} }  -1  + \epsilon \right | \\
    =  \left | \left(\frac{\rm{Vol}_q^c}{\rm{Vol}_q^{*}}\right)^2 Z_q -1 \right | 
    =  \left |  R_q^{2} Z_q -1 \right |,
    %=\log \mathcal{P}_q (M_q|X) =\log \det(L_{M_q}(X;q,\theta)) - \log \det(L(X;q,\theta)+I)).
\end{split}
\end{equation}
%where $Z_q=\frac{\sum_{|\widetilde{M}_q^*|=N_q^{mb}} \det(L_{\widetilde{M}_q^*}(D;  \theta))}{\sum_{|\widetilde{M}_q^c|=N_q^{mb}} \det(L_{\widetilde{M}_q^c}(D;  q, \theta))}$ is the ratio of the two denominator determinants. Similar to the geometric interpretation of \(\det(M_q^{* \top} M_q^{*})\), \(\det(M_q^{c\top} M_q^c) = ({\rm{Vol}_q^c})^2\) is the squared volume of \(M_q^c\). Specifically, $ \rm{Vol}_q^c =  \rm{Vol}(\rm{Conv}({\bar{0}}\cup M_q^c))$ and $ \rm{Vol}_q^* =  \rm{Vol}(\rm{Conv}({\bar{0}}\cup M_q^*))$. We define \(R_q = \frac{\rm{Vol}_q^c}{\rm{Vol}_q^{*}}\) as the ratio of the two feature volumes. In essence, we can convert optimizing \(\mathcal{L}_2 \) into minimizing \(|R_q - 1|\), so as to avoid the huge computation cost of \(Z_q\) (detailed in Proposition 2).
where $Z_q=\frac{\sum_{|M^*|=N_q^{mb}} \det(L_{M^*}(D;  \theta))}{\sum_{|M|=N_q^{mb}} \det(L_{M}(D;  q, \theta))}$. In particular, $\det(M_q^{* \top} M_q^*)$ has a geometric interpretation that it is equal to the squared volume spanned by $M_q^*$ \citep{kulesza2012determinantal}, denoted as $\rm{Vol}_q^{*}$, likewise for $\det(M_q^{c \top} M_q^c)$ with respect to $\rm{Vol}_q^{c}$. Then we define \(R_q = \frac{\rm{Vol}_q^c}{\rm{Vol}_q^{*}}\) as the ratio of the two feature volumes. To avoid computing \(Z_q\), we can convert optimizing \(\mathcal{L}_2 \) into minimizing \(|R_q - 1| \) equivalently, since both of them mean maximizing $q$ (detailed in Proposition~\ref{propo2} as below).

% \begin{lemma}
% For any image dataset $D=(x_1,x_2,\cdots,x_n)$, where $x_i$ is the $i$-th image of $D$, define $D_q=(x_1,x_2,\cdots,x_n)$ as a compressed image dataset with the compression ratio $q$ for $D$.
% Additionally, denote
% $X=({\bm{f}_1,\bm{f}_2,\cdots,\bm{f}_n}) \in \mathcal{R}^{d \times n}$ and  $X_q=({\bm{f}_1^q,\bm{f}_2^q,\cdots,\bm{f}_n^q}) \in \mathcal{R}^{d \times n}$ as the extracted feature matrices of $D$ and $D_q$, respectively.
% Further, define $X_M=({\bm{f}_{i_i},\bm{f}_{i_2},\cdots,\bm{f}_{i_{m}}})\in \mathcal{R}^{d \times m}$ as a submatrix sampled from $X$ with the cardinality $|X_M|=s$, and $X_M^q=({\bm{f}_{i_i}^q,\bm{f}_{i_2}^q,\cdots,\bm{f}_{i_{m}}^q})\in \mathcal{R}^{d \times m}$ as a submatrix sampled from $X_q$ with the cardinality $|X_M^q|=m$.
% Then for 
% \begin{align*}
% R_q := \sqrt{\frac{\det(M_q^{\rm T} M_q)}{\det(M_q^{*T} M_q^{*})}},
% \end{align*}
% % and
% % \begin{align*}
% % R_q := \sqrt{\frac{\det(M_q^{\rm T} M_q)}{\det(M_q^{*T} M_q^{*})}},
% % \end{align*}
% we can ensure that
% \begin{align*}
% q^* = 100 = \text{argmin}_{q} \ |R_q - 1|.
% \end{align*}
% \end{lemma}
% This means when $q=100$, the $|R_q - 1|$ takes the minimum value.

Below we are going to introduce Proposition~\ref{propo2}. We first introduce some necessary assumptions, then we present the theoretical guarantee of Proposition~\ref{propo2}.

\begin{assumption}\label{assume1}
We make the following assumptions:
\begin{enumerate}
    \item \textbf{Image dataset:} For any image dataset $D=(x_1,x_2,\cdots,x_n)$, where $x_i$ is the $i$-th image in $D$, define $D_q=(x_1,x_2,\cdots,x_n)$ as a compressed image dataset with the compression quality $q$ for $D$. The compression quality is bounded as $q \in Q$ (Q can be a finite/infinite set), in which the higher of $q$, the better of image quality.
    \item \textbf{Feature matrix:} Denote $X=({\bm{f}_1,\bm{f}_2,\cdots,\bm{f}_n}) \in \mathcal{R}^{d \times n}$ and  $X_q=({\bm{f}_1^q,\bm{f}_2^q,\cdots,\bm{f}_n^q}) \in \mathcal{R}^{d \times n}$ as the feature matrices of $D$ and $D_q$ with the same fixed feature extractor, respectively. Here, $d > n$, $X^{\top}X\neq I$ and $X_q^{\top} X_q \neq I$.
    \item \textbf{Selection principle:} Define $\mathcal{M}=\{i_1,i_2,\cdots,i_m\}$ as a subset of $[n]$ containing $m$ elements, and $X_{\mathcal{M}}$ as the submatrix sampled from $X$ using indices from $\mathcal{M}$.
    Likewise, define $\mathcal{M}_{q}=\{j_1,j_2,\cdots,j_m\}$ as a subset of $[n]$ containing $m$ elements, and $X_{\mathcal{M}_q}$ as the submatrix sampled from $X_q$ using indices from $\mathcal{M}_{q}$. Both of them use the same selection principle.
    % We make the assumption that $X_{\mathcal{M}}$ (resp., $X_{\mathcal{M}^q}$) is selected from $\binom n m$ possible options in $X$ (resp., $X_q$). 
    \item \textbf{Buffer storage space:} $g(\cdot): \mathbb{R} \rightarrow \mathbb{N}$ represents the function that outputs the maximum number (i.e. \(m\) ) such that the compressed version of \(D_q\) to a quality \(q\) can be stored in the memory buffer. That is, $m$ is uniquely determined by $q$.
\end{enumerate}
\end{assumption}

\begin{proposition} \label{propo2}
Under Assumption~\ref{assume1}, for two probabilities
%  \begin{align*}
%  R_q := \frac{\det\left(X_{\mathcal{M}^q}^{\top}X_{\mathcal{M}^q}\right)}{\det\left(X_{\mathcal{M}}^{\top}X_{\mathcal{M}}\right)},
%  \end{align*}
%  and 
\begin{align*}
P(\mathcal{M}):= \frac{\det\left(X_{\mathcal{M}}^{\top}X_{\mathcal{M}}\right)}{\sum_{|{\mathcal{{M}}'}|=m} \ \det\left({X}_{{\mathcal{{M}}'}}^{\top}{X}_{{\mathcal{{M}}'}}\right)}
\end{align*}
and 
\begin{align*}
P({\mathcal{M}_q}):= \frac{\det\left(X_{M_q}^{\top}X_{M_q}\right)}{\sum_{|\widetilde{\mathcal{M}}|=m} \ \det\left({X}_{\widetilde{\mathcal{M}}}^{\top}{X}_{\widetilde{\mathcal{M}}}\right)},
\end{align*}
% and
% \begin{align*}
% R_q := \sqrt{\frac{\det(M_q^{\rm T} M_q)}{\det(M_q^{*T} M_q^{*})}},
% \end{align*}
we can ensure that the following two optimization programs 
\begin{align*}
    \min_q \ \left|\frac{\det\left(X_{\mathcal{M}_q}^{\top}X_{\mathcal{M}_q}\right)}{\det\left(X_{\mathcal{M}}^{\top}X_{\mathcal{M}}\right)} - 1 \right|
\end{align*}
and
\begin{align*}
    \min_q \ \left|\frac{ P(\mathcal{M}_q) }{ P({\mathcal{M}})} - 1 \right|
\end{align*}
are equivalent.
\end{proposition}

Below are two critical lemmas for the proof of Proposition~\ref{propo2}.

\begin{lemma}\label{lemma2}
Under Assumption~\ref{assume1}, for a ratio
 \begin{align*}
 R_q := \frac{\det\left(X_{\mathcal{M}_q}^{\top}X_{\mathcal{M}_q}\right)}{\det\left(X_{\mathcal{M}}^{\top}X_{\mathcal{M}}\right)},
 \end{align*}
we can ensure that
\begin{align*}
\sup Q = \text{argmin}_{q} \ |R_q - 1|.
\end{align*}
This means when $q= \sup Q$ (i.e, without image compression), $|R_q - 1|$ takes the minimum value.
\end{lemma}

\noindent\emph{Proof.}
From the definition, we know $|R_q - 1| \geq 0$.
To minimize $|R_q - 1|$, we need the ratio $R_q=1$, i.e.,
\begin{align} \label{lemma2-equ}
    \det\left(X_{\mathcal{M}_q}^{\top}X_{\mathcal{M}_q}\right)=\det\left(X_{\mathcal{M}}^{\top}X_{\mathcal{M}}\right).
\end{align}
Of note, both $X_{\mathcal{M}_q}$ and $X_{\mathcal{M}}$ are selected from the source dataset $D$ with the same feature extractor, selection principle and set cardinality (i.e., $m$). That is, the only difference between them is image compression with respect to $q$.

By recurring to the Geometric Interpretation in $k$-DPPs (i.e., Eq.~(\ref{kdpp-vol})), we have
\begin{align*}
     \det\left(X_{\mathcal{M}_q}^{\top}X_{\mathcal{M}_q}\right) = 
    \ (m!\cdot (\rm{Vol}(\rm{Conv({\bar{0}\cup {X_{\mathcal{M}_q}}}})))^2)
\end{align*}
and 
\begin{align*}
     \det\left(X_{\mathcal{M}}^{\top}X_{\mathcal{M}}\right) = 
    \ (m!\cdot (\rm{Vol}(\rm{Conv({\bar{0}\cup {X_{\mathcal{M}}}}})))^2).
\end{align*}
Then, Eq.~(\ref{lemma2-equ}) holds only when $X_{\mathcal{M}_q}$ and $X_{\mathcal{M}}$ have the same convex hull.
Due to great difficulty of analytically defining compression function, the two convex hull cannot be mathematically given.
However, by conducting extensive experiments on this task (as shown in Fig.~\ref{Feature_Manifold_Compressed}, Fig.~\ref{5phase_ImageNet_Rq}, Fig.~\ref{SODA10M_Rq}, Fig.~\ref{Volume_by_Phase_All} and Fig.~\ref{Cumu_Volume_AANets_DDE}), we find that the volume of $X_{\mathcal{M}_q}$ is larger than that of $X_{\mathcal{M}}$ for a specific $m$ (i.e., $q$).
Additionally, with $q$ decreasing (i.e., $m$ increasing), the volume of $X_{\mathcal{M}_q}$ is increasing more quickly. Thus, we have an empirical conclusion that the two volumes are the same only when $q =\sup Q$ (without image compression). It is reasonable since without image compression, the two selection problems about $X_{\mathcal{M}_q}$ and $X_{\mathcal{M}}$ are identical.
This means when $q= \sup Q$, $|R_q - 1|$ takes the minimum value, which completes the proof.

\begin{lemma}\label{lemma3}
Under Assumption~\ref{assume1}, for 
\begin{align*}
P(\mathcal{M}):= \frac{\det\left(X_{\mathcal{M}}^{\top}X_{\mathcal{M}}\right)}{\sum_{|{\mathcal{{M}}'}|=m} \ \det\left({X}_{{\mathcal{{M}}'}}^{\top}{X}_{{\mathcal{{M}}'}}\right)}
\end{align*}
and 
\begin{align*}
P({\mathcal{M}_q}):= \frac{\det\left(X_{M_q}^{\top}X_{M_q}\right)}{\sum_{|\widetilde{\mathcal{M}}|=m} \ \det\left({X}_{\widetilde{\mathcal{M}}}^{\top}{X}_{\widetilde{\mathcal{M}}}\right)},
\end{align*}
we can ensure that
\begin{align*}
\sup Q = \text{argmin}_{q}  \left |  \frac{ P(\mathcal{M}_q) }{ P({\mathcal{M}})} -1 \right | .
\end{align*}
This means when $q=\sup Q$ (i.e, without image compression), $\left |  \frac{ P(\mathcal{M}_q) }{ P({\mathcal{M}})} -1 \right |$ takes the minimum value.
\end{lemma}

\noindent\emph{Proof.}
From the definition, we know $\left |  \frac{ P(\mathcal{M}_q) }{ P({\mathcal{M}})} -1 \right |\geq 0$.
To minimize $\left |  \frac{ P(\mathcal{M}_q) }{ P({\mathcal{M}})} -1 \right |$, we need the ratio $\frac{ P(\mathcal{M}_q) }{ P({\mathcal{M}})}=1$.
In fact,
\begin{align}
     \frac{ P(\mathcal{M}_q) }{ P({\mathcal{M}})} &= \frac{\det\left(X_{M_q}^{\top}X_{M_q}\right)}{\det\left(X_{\mathcal{M}}^{\top}X_{\mathcal{M}}\right)}\cdot
     \frac{\sum_{|{\mathcal{{M}}'}|=m} \ \det\left({X}_{{\mathcal{{M}}'}}^{\top}{X}_{{\mathcal{{M}}'}}\right)}
     {\sum_{|\widetilde{\mathcal{M}}|=m} \ \det\left({X}_{\widetilde{\mathcal{M}}}^{\top}{X}_{\widetilde{\mathcal{M}}}\right)}  = R_q\cdot Z_q,
\end{align}
where $R_q = \frac{\det\left(X_{\mathcal{M}_q}^{\top}X_{\mathcal{M}_q}\right)}{\det\left(X_{\mathcal{M}}^{\top}X_{\mathcal{M}}\right)}$ and $Z_q = \frac{\sum_{|{\mathcal{{M}}'}|=m} \ \det\left({X}_{{\mathcal{{M}}'}}^{\top}{X}_{{\mathcal{{M}}'}}\right)}
     {\sum_{|\widetilde{\mathcal{M}}|=m} \ \det\left({X}_{\widetilde{\mathcal{M}}}^{\top}{X}_{\widetilde{\mathcal{M}}}\right)}$.
     
    By using Lemma~\ref{lemma2}, we know that when $q=\sup Q$, $R_q = 1$ always holds. More importantly, $Z_q = 1$ also holds if $q=\sup Q$, since the numerator and denominator in $Z_q$ will be equal without compression.
    That is, when $q=\sup Q$, the ratio $\frac{ P(\mathcal{M}_q) }{ P({\mathcal{M}})}=1$ and $\left |  \frac{ P(\mathcal{M}_q) }{ P({\mathcal{M}})} -1 \right |$ takes the minimum value.
    This completes the proof.

\noindent \textbf{Proof of Proposition~\ref{propo2}} From Lemma~\ref{lemma2} and Lemma~\ref{lemma3}, we can ensure that under Assumption~\ref{assume1}, the following two optimization programs 
\begin{align*}
    \min_q \ \left|\frac{\det\left(X_{\mathcal{M}_q}^{\top}X_{\mathcal{M}_q}\right)}{\det\left(X_{\mathcal{M}}^{\top}X_{\mathcal{M}}\right)} - 1 \right|
\end{align*}
and
\begin{align*}
    \min_q \ \left|\frac{ P(\mathcal{M}_q) }{ P({\mathcal{M}})} - 1 \right|
\end{align*}
have the same optimal value.
Thus, maximizing $ \left|\frac{ P(\mathcal{M}_q) }{ P({\mathcal{M}})} - 1 \right|$ with respect to $q$ can be converted into $\min_{q} |R_q - 1|$ equivalently, where $R_q=\frac{\det\left(X_{\mathcal{M}_q}^{\top}X_{\mathcal{M}_q}\right)}{\det\left(X_{\mathcal{M}}^{\top}X_{\mathcal{M}}\right)}$.

\section{Empirical Analysis}

\subsection{t-SNE Visualization of Normalized Features}
\begin{figure}[th]
    \centering
    \includegraphics[width=0.95\linewidth]{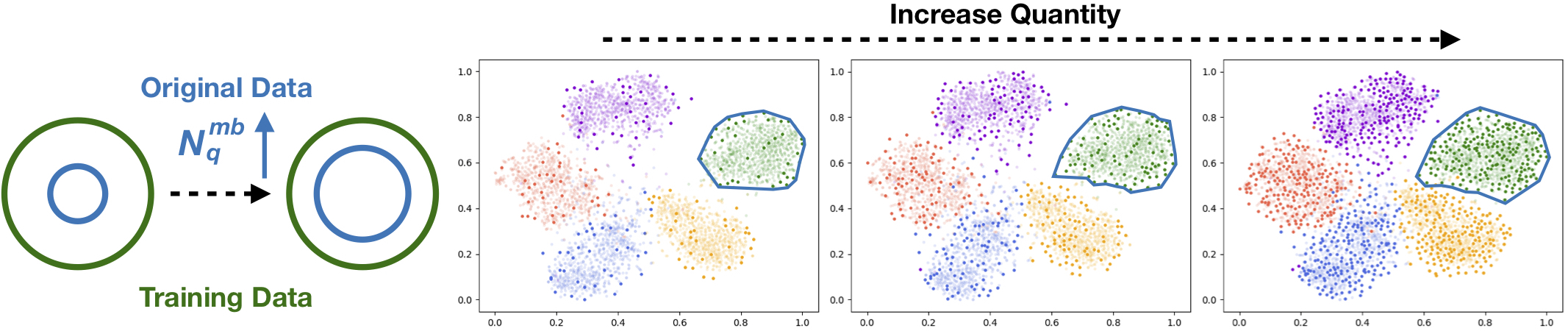}
    \caption{t-SNE visualization of features of original subsets (dark dots) and all training data (light dots) for 5-phase ImageNet-sub with LUCIR. We randomly select five classes out of the latest task and label them in different colors.} 
    \label{Feature_Manifold_Original}
    %\vspace{-0.1cm}
\end{figure}

To provide an empirical analysis of the quality-quantity trade-off and validate the theoretical interpretation, we use t-SNE \citep{van2008visualizing} to visualize features of all training data, the compressed subset \(M_q\), and the original subset \(M_q^*\). 
%First, if the quality is high enough, larger quantity will improve memory replay via better recovering the old data distribution. 
First, with the increase of quantity \(N_q^{mb}\), the area of original subset is expanded and can better cover the training data distribution, as shown in Fig. \ref{Feature_Manifold_Original}. This result is consistent with maximizing \(N_q^{mb}\) for $\mathcal{L}_1$.
Second, with the decrease of quality $q$ and increase of quantity \(N_q^{mb}\), the compressed data tend to be distorted and thus become out-of-distribution, which has been discussed in the main text Fig. \ref{Feature_Manifold_Compressed}. This result is consistent with enforcing \(|R_q - 1| < \epsilon\) for $\mathcal{L}_2$.

%Similarly, we visualize the coverage of compressed data on the same amount of original data in Fig. \ref{Feature_Manifold} (b). With the decrease of quality and increase of quantity, the area of compressed data is initially similar to that of original data, then they expand synchronously, but as numerous low-quality compressed data occur out-of-distribution, the area of compressed data becomes much larger than that of original data. This leads to performance declining in continual learning as shown in Fig. \ref{Accuracy_Tradeoff}.

\subsection{\(R_q\) for 5-, 10- and 25-phase ImageNet-sub}

\begin{figure}[th]
    \centering
    \includegraphics[width=0.95\linewidth]{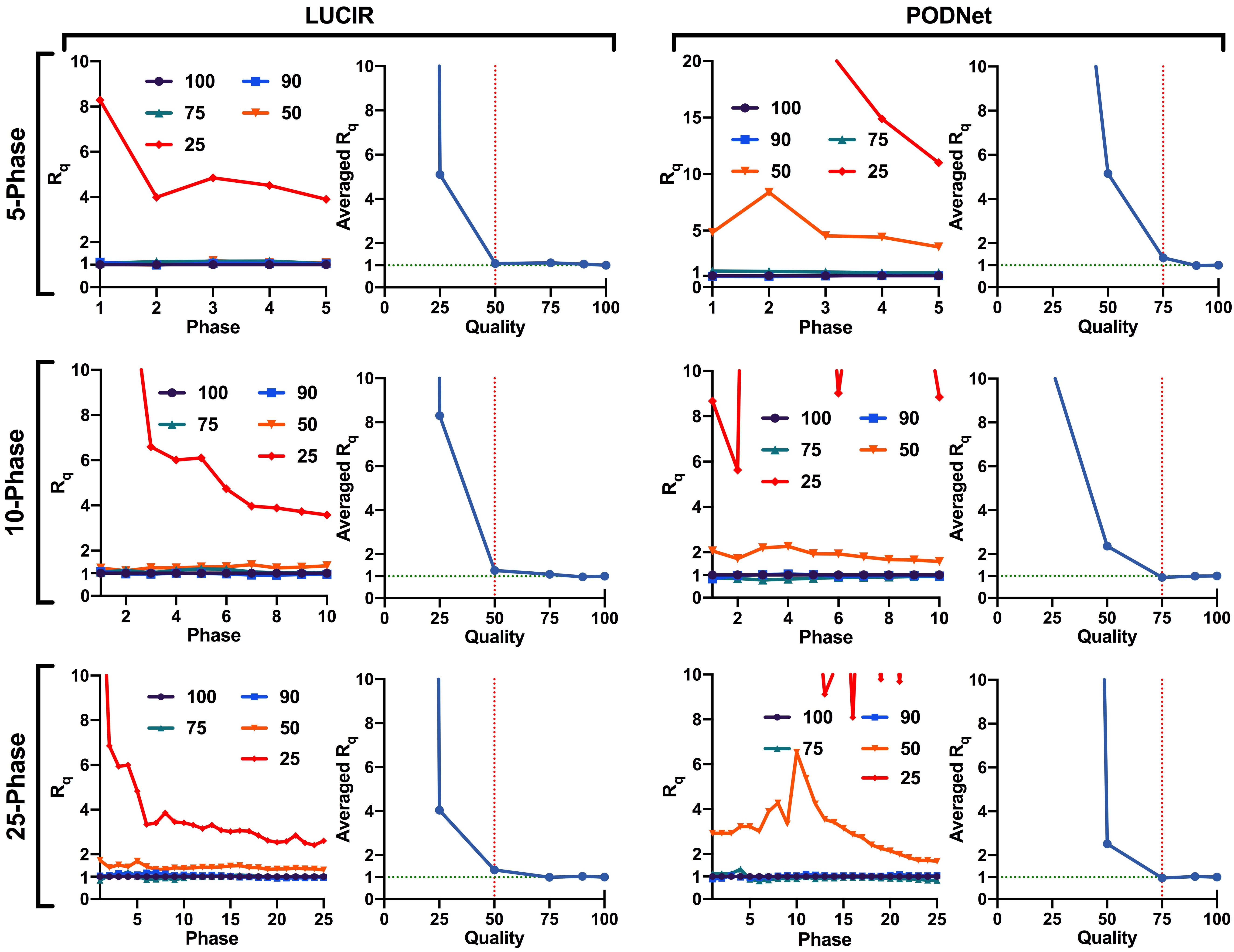}
    \vspace{-0.2cm}
    \caption{We present \(R_q\) in each incremental phase with various compression qualities (left), and the averaged \(R_q\) of all incremental phases (right). From top to bottom are 5-, 10- and 25-phase ImageNet-Sub, respectively.}
    \label{Volume_by_Phase_All}
    \vspace{-0.3cm}
\end{figure}

We present \(R_q\) in each incremental phase and the averaged \(R_q\) of all incremental phases for 5-, 10- and 25-phase ImageNet-sub in Fig.\ref{Volume_by_Phase_All}. Based on the principle in Eq.~(\ref{Objective}) (we set \(\epsilon = 0.5\) as the threshold of \(R_q\)), it can be clearly seen that 50 and 75 is a good quality for LUCIR and PODNet, respectively. Also, the determined quality is the same for different numbers of splits, which is consistent with the results of grid search in Fig. \ref{Accuracy_Tradeoff}.

%Here we present the phase-by-phase plot of \(R_q\) on 5-, 10- and 25-phase split ImageNet-sub with LUCIR \citep{hou2019learning} and PODNet \citep{douillard2020podnet} in Fig. \ref{Volume_by_Phase_All}. 50 and 75 is the smallest quality that satisfies \(|R_q - 1| < \epsilon\) (we empirically choose \(\epsilon=0.5\)) for LUCIR and PODNet in most of the incremental phases, respectively, although \(R_q\) slightly varies in the 25-phase split. 

\subsection{AANets and DDE}
\begin{figure}[th]
    \centering
    \includegraphics[width=0.95\linewidth]{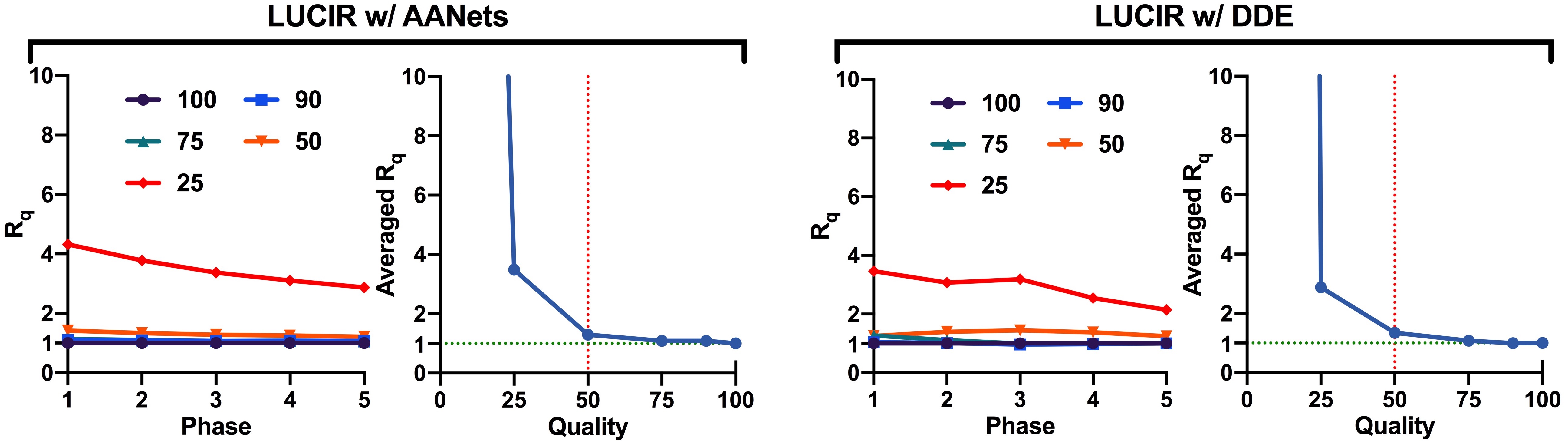}
    \vspace{-0.2cm}
    \caption{Determine the compression quality for AANets and DDE. We present \(R_q\) in each incremental phase (left), and the averaged \(R_q\) of all incremental phases (right) for 5-phase ImageNet-sub. The quality of 100 refers to the original data without compression.}
    \label{Cumu_Volume_AANets_DDE}
    \vspace{-0.2cm}
\end{figure}

Similar to LUCIR and PODNet, we apply the method described in Sec.~4.2 to determine the compression quality for AANets and DDE. Since both AANets and DDE only release their official implementation on LUCIR, here we focus on LUCIR w/ AANets and LUCIR w/ DDE. 
We present \(R_q\) in each incremental phase and the averaged \(R_q\) of all incremental phases for 5-phase ImageNet-sub in Fig. \ref{Cumu_Volume_AANets_DDE}. 
The determined qualities for both LUCIR w/ AANets and LUCIR w/ DDE are consistent in different incremental phases, and are the same as that of LUCIR.

\subsection{Static vs Dynamic Quality}
For ImageNet-sub with randomly-split classes, whether \(|R_q - 1| < \epsilon\) of each quality \(q\) is consistent among incremental phases and their average (see Fig.~\ref{Volume_by_Phase_All} and Fig.~\ref{Cumu_Volume_AANets_DDE}). Thus, the determined quality by our method usually serves as a static hyperparameter in continual learning, which has been validated by the results of gird search (see Fig. \ref{Accuracy_Tradeoff}).
%As the determined quality by our method has been validated by the results of gird search (see Fig. \ref{Accuracy_Tradeoff}), it serves as a static hyperparameter in continual learning. 
Now, we wonder in which case the determined quality is dynamic, and whether using a dynamic quality is better than using a static one. Intuitively, the determined quality might be affected by the similarity of incremental classes, because the model learned to predict similar classes might be more sensitive to the image distortion caused by data compression. 

%\begin{figure}[th]
\begin{wrapfigure}{r}{0.60\textwidth}
    \centering
    \vspace{-0.3cm}
    \includegraphics[width=1\linewidth]{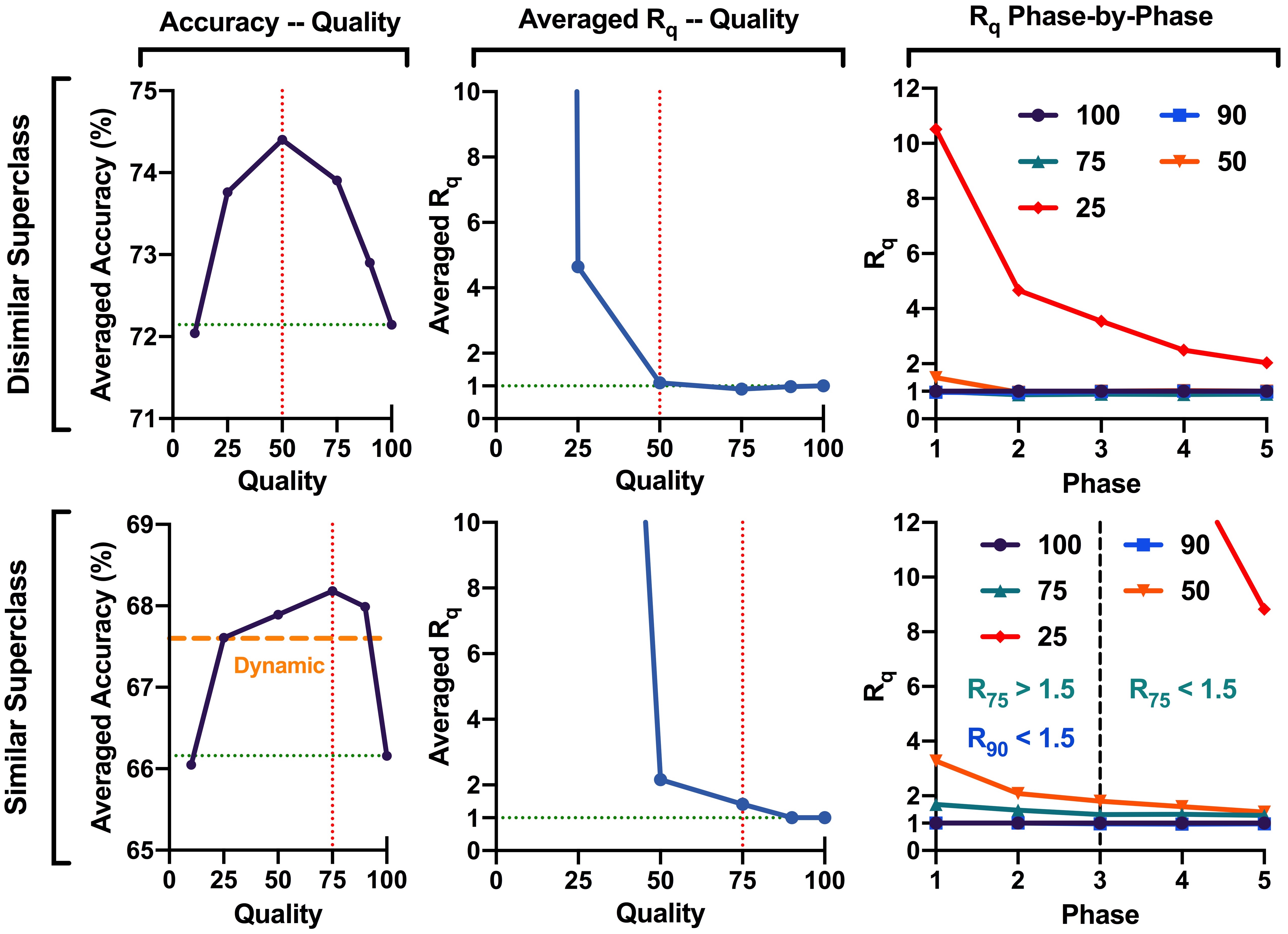}
    \vspace{-0.7cm}
    \caption{Continual learning of ten similar or dissimilar superclasses with LUCIR. } 
     \vspace{-0.3cm}
    \label{Accuracy_Volume_SC}
    %\vspace{-0.5cm}
\end{wrapfigure}
%\end{figure}

Since the ImageNet classes are ranked by their semantic similarity, we select 10 adjacent classes as a superclass, and construct 10 adjacent (similar) or remote (dissimilar) superclasses (detailed in Appendix B.1). Similar to the 5-phase ImageNet-sub, we first learn 5 superlcasses as the initial phase, and then incrementally learn one superclass per phase with LUCIR. The results are presented in Fig. \ref{Accuracy_Volume_SC}. 
For dissimilar superclasses, whether \(|R_q - 1| < \epsilon\) is consistent in each incremental phase and their average, so the determined quality is stable at 50, validated by the results of grid search. 
For similar superclasses, the quality determined by the averaged \(R_q\) is 75, where using a stable quality of 75 indeed achieves the best performance in grid search. On the other hand, it can be clearly seen that \(|R_{75} - 1| > \epsilon\) before phase 3 while \(|R_{75} - 1| < \epsilon\) after it. Then we use the dynamically-determined quantity and quantity for each incremental phase. However, the dynamic quality and quantity result in severe data imbalance, so the performance is far lower than using a static quality of either 50, 75 or 90. A promising further work is to alleviate the data imbalance for the scenarios where the determined quality is highly dynamic.

\section{Additional Results}

\subsection{Computational Cost}
In Table \ref{table:time_cost}, we present the detailed results of computational cost and averaged incremental accuracy of LUCIR, LUCIR w/ AANets, LUCIR w/ DDE and LUCIR w/ MRDC for CUB-200-2011 and ImageNet-sub. %In particular, AANets needs to learn additional parameters, resulting in more storage costs in addition to the memory buffer. %We further plot the results on CUB-200-2011 in Fig. \ref{Time+Accuracy_CUB}.

\begin{table*}[th]
	\centering
    \caption{Comparison of computational cost and averaged incremental accuracy. We run each baseline with one Tesla V100.}
    %\vspace{-.1cm}
	\smallskip
	\resizebox{1\textwidth}{!}{ 
	\begin{tabular}{lcccccc}
		\specialrule{0.01em}{1.2pt}{1.5pt}
		 \multicolumn{1}{c}{} & \multicolumn{3}{c}{CUB-200-2011} & \multicolumn{3}{c}{ImageNet-sub} \\
      \cmidrule(lllr){2-4}  \cmidrule(lllr){5-7}
       Methods & 5-phase & 10-phase & 25-phase & 5-phase & 10-phase & 25-phase \\
    \specialrule{0.01em}{1.2pt}{1.7pt}
    LUCIR &1.97 h / 44.63\% &3.60 h / 45.58\%&8.03 h / 45.48\%&8.65 h / 70.84\%&10.47 h / 68.32\%&15.63 h / 61.44\% \\
    %PODNet &4.38 h / \% &7.21 h / \%&17.09 h / \%&47.19 h / 75.54\%&52.96 h / 74.33\%&80.35 h / 68.31\% \\
    LUCIR w/ AANets &3.52 h / 46.87\%&5.56 h / 47.34\%&14.34 h / 47.35\% &21.94 h / 72.55\% &40.58 h / 69.22\% &91.26 h / 67.60\% \\
    LUCIR w/ DDE &7.28 h / 45.86\% &13.09 h / 46.48\% &31.02 h / 46.56\% &55.36 h / 72.34\% &61.81 h / 70.20\% &79.42 h / 66.31\% \\
    LUCIR w/ \emph{MRDC (Ours)} &2.70 h / 46.68\% &5.00 h / 47.28\% &11.00 h / 48.01\% &9.09 h / 73.56\% &11.27 h / 72.70\% &18.83 h / 70.53\% \\
    \specialrule{0.01em}{1.2pt}{1.7pt}
	\end{tabular}
	}
	\label{table:time_cost}
	%\vspace{-.1cm}
\end{table*}

\subsection{Storage Space}
In Table \ref{table:buffer_size}, we present the detailed results of different storage space. 

\begin{table}[th]
    \vspace{-.3cm}
	\centering
    \caption{Averaged incremental accuracy (\(\%\)) on ImageNet-sub. The storage space of the memory buffer is limited to the equivalent of 10, 20, 40 and 80 original images per class, respectively. The results of LUCIR and PODNet are reproduced from their officially-released codes.}
    %\vspace{-.1cm}
	\smallskip
	\resizebox{0.75\textwidth}{!}{ 
	\begin{tabular}{clcccc}
		\specialrule{0.01em}{1.2pt}{1.5pt}
      & Storage Space & 10 & 20 & 40 & 80 \\
      \specialrule{0.01em}{1.2pt}{1.7pt}
      \multirow{4}*{\tabincell{c}{5-phase}}
      & LUCIR  &68.83 &70.90 &72.64 &73.56 \\
      & w/ \emph{MRDC (Ours)} &71.39 / +2.55 &73.56 / +2.66 &74.37 / +1.73 &75.74 / +2.19 \\
      \cdashline{2-6}[4pt/4pt]
      & PODNet &74.99 &76.44 &78.28 &79.18 \\
      & w/ \emph{MRDC (Ours)} &76.55 / +1.56 &78.08 / +1.64 &79.43 / +1.15 & 80.17 / +0.99 \\
     \specialrule{0.01em}{1.2pt}{1.7pt}
      \multirow{4}*{\tabincell{c}{10-phase}}
      & LUCIR &64.64 &68.31 &70.84 &72.56 \\
      & w/ \emph{MRDC (Ours)} &68.13 / +3.49 &72.70 / +4.39 &74.11 / +3.28 &76.09 / +3.54 \\
      \cdashline{2-6}[4pt/4pt]
      & PODNet &70.73 &73.91 &76.63 &78.26 \\
      & w/ \emph{MRDC (Ours)} &72.78 / +2.05 &76.02 / +2.11 & 77.82 / +1.19 &79.27 / +1.01 \\
     \specialrule{0.01em}{1.2pt}{1.7pt}
      \multirow{4}*{\tabincell{c}{25-phase}}
      & LUCIR &58.98 &64.46 &67.55 &70.12 \\
      & w/ \emph{MRDC (Ours)} &67.75 / +5.77 &70.53 / +6.07 &73.49 / +5.94 &76.03 / +5.91 \\
      \cdashline{2-6}[4pt/4pt]
      & PODNet &59.41 &67.17 &72.57 &76.21 \\
      & w/ \emph{MRDC (Ours)} &64.99 / +5.58 &72.27 / +5.10 &76.61 / +4.04 &78.82 / +2.61 \\
     \specialrule{0.01em}{1.2pt}{1.7pt}
	\end{tabular}
	}
	\label{table:buffer_size}
	%\vspace{-.1cm}
\end{table}

\subsection{Averaged Forgetting}
In addition to averaged incremental accuracy, we evaluate the averaged forgetting, which is calculated by averaging the test accuracy of each task minus its highest test accuracy achieved in continual learning \citep{caccia2020online}. In Table \ref{table:averaged_forgetting}, we present the averaged forgetting of classes in the initial phase, which suffers from the most severe forgetting. It can be clearly seen that the averaged forgetting is largely alleviated by ours on each backbone.

\renewcommand\arraystretch{1.5}
\begin{table*}[ht]
	\centering
    \vspace{-.4cm}
    \caption{Averaged forgetting (\(\%\)) of classes in the initial phase. The storage space is limited to the equivalent of 20 original images per class. DDE \citep{hu2021distilling} and AANets \citep{liu2021adaptive} are reproduced from their officially-released code.}
    %For CUB-200-2011, the performance of all baselines is reproduced from their officially-released code (if applicable), while for ImageNet-sub and ImageNet-full we present the reported performance of all baselines and the reproduced results of AANets and DDE, so as to make the comparison as fair as possible. The reproduced results might slightly vary from the reported results due to the usage of different random seeds. \(^1\)With class-balance finetuning \citep{hu2021distilling}. \(^2\)PODNet reproduced by \cite{hu2021distilling} underperforms that in \cite{douillard2020podnet}. 
    %\vspace{-.1cm}
	\smallskip
	\resizebox{0.7\textwidth}{!}{ 
	\begin{tabular}{lcccccc}
		\specialrule{0.01em}{1.2pt}{1.5pt}
		 \multicolumn{1}{c}{} & \multicolumn{3}{c}{CUB-200-2011} & \multicolumn{3}{c}{ImageNet-sub} \\
    \cmidrule(lllr){2-4}  \cmidrule(lllr){5-7}
       Method & 5-phase & 10-phase & 25-phase & 5-phase & 10-phase & 25-phase \\
       \specialrule{0.01em}{1.2pt}{1.7pt}
       LUCIR \citep{hou2019learning}& -2.86 & -4.56 & -4.72 & -15.08 & -17.32 & -22.40\\
      \rowcolor{black!15}
      \quad w/ \emph{MRDC (Ours)}& -0.84 & -1.08 & -0.75 & -14.11 & -14.64 & -17.96\\
      \cdashline{1-7}[2pt/2pt]
     \quad w/ AANets (Reproduced) & -4.03 & -6.58 & -10.37 & -11.78 & -9.54 & -12.86 \\
       \rowcolor{black!15}
      \quad w/ AANets + \emph{MRDC (Ours)}& -1.40 & -3.77 & -6.82 & -7.46 & -8.13 & -12.32\\
      \cdashline{1-7}[2pt/2pt]
       \quad w/ DDE (Reproduced) & -1.41 & -5.24 & -3.91 & -14.02 & -17.68 & -23.73\\
        \rowcolor{black!15}
      \quad w/ DDE + \emph{MRDC (Ours)} & -0.22 & -0.97 & -0.71 & -11.24 & -12.36 & -17.80\\
    \specialrule{0.01em}{1.2pt}{1.7pt}
	\end{tabular}
	}
	\label{table:averaged_forgetting}
	\vspace{-.2cm}
\end{table*}

\subsection{Fixed Memory Budget}

Here we evaluate memory replay with data compression (MRDC, ours) under a fixed memory budget (i.e., storage space) on ImageNet-sub. Following a widely-used setting \citep{rebuffi2017icarl,wu2019large}, the memory budget (i.e., storage space) is limited to equivalent of 2000 original images (20 original images per class $\times$ a total of 100 classes in ImageNet-sub). We further evaluate a much smaller memory budget, fixed to 1000 original images.
Under such a fixed memory budget, ours substantially boosts the performance of each backbone approach as shown in Table \ref{table:fixed_memory_budget}. 
 
\renewcommand\arraystretch{1.5}
\begin{table*}[th]
	\centering
    \vspace{-.3cm}
    \caption{Averaged incremental accuracy (\%) on ImageNet-sub, under a fixed memory budget of 1000 or 2000 original images. }
	\smallskip
	\resizebox{0.7\textwidth}{!}{ 
	\begin{tabular}{lcccccc}
		\specialrule{0.01em}{1.2pt}{1.5pt}
		 \multicolumn{1}{c}{} & \multicolumn{3}{c}{1000 Original Images} & \multicolumn{3}{c}{2000 Original Images}\\
    \cmidrule(lllr){2-4}  \cmidrule(lllr){5-7}  
       Method & 5-phase & 10-phase & 25-phase & 5-phase & 10-phase & 25-phase \\
       \specialrule{0.01em}{1.2pt}{1.7pt}
       LUCIR \citep{hou2019learning}&70.16 & 66.32 & 62.07 & 71.87 &69.37 &65.72 \\
      \rowcolor{black!15}
      \quad w/ \emph{MRDC (Ours)}&72.37 &69.63 &68.16 & 73.73 &73.07 &72.02 \\
      %\cdashline{1-4}[2pt/2pt]
     %\quad w/ AANets (Reproduced)&  &  & \\
      % \rowcolor{black!15}
     % \quad w/ AANets + \emph{Ours}&  &  & \\
      \cdashline{1-4}[2pt/2pt]
       \quad w/ DDE (Reproduced) &72.61 &70.82 &65.58 & 73.47  &71.92  &66.70 \\
        \rowcolor{black!15}
      \quad w/ DDE + \emph{MRDC (Ours)}&74.40 &72.62 &69.03 & 75.35  &73.93  &71.17 \\
    \specialrule{0.01em}{1.2pt}{1.7pt}
	\end{tabular}
	}
	\label{table:fixed_memory_budget}
	\vspace{-.2cm}
\end{table*}

\subsection{Less Compressed Samples}

In Table \ref{table:less_compressed_samples}, we present the results of LUCIR with different numbers of compressed samples on ImageNet-sub. Similar to the main text, we select the JPEG quality of $50$ for LUCIR, where the storage space of $20$ original images can save $85$ such compressed images. Memory replay of $85$ compressed images of quality $50$ achieves a much better performance than that of $20$ original images. When reducing the quantity from $85$ to $70$, $55$ and $40$, the accuracy will also decline. However, memory replay of $40$ compressed images of quality $50$ still outperforms that of $20$ original images, where the average memory can be saved by $52.94\%$.

\renewcommand\arraystretch{1.5}
\begin{table*}[th]
	\centering
    \vspace{-.2cm}
    \caption{Averaged incremental accuracy (\%) of LUCIR on ImageNet-sub with different numbers of compressed data. ``5-phase'', ``10-phase'' and ``25-phase'' refer to the accuracy of 5-, 10- and 25-phase ImageNet-sub, respectively.}
	\smallskip
	\resizebox{0.6\textwidth}{!}{ 
	\begin{tabular}{cccccc}
		\specialrule{0.01em}{1.2pt}{1.5pt}
       Quality & 50 & 50 & 50 & 50 & Original \\
       Quantity &85 &70 &55 &40 &20 \\
      Total Storage &100\% &82.35\% &64.71\% &47.06\% &100\% \\
      \cdashline{1-6}[2pt/2pt]
      5-phase  &73.63 &73.22 &72.79 &72.29 &72.06 \\
     10-phase &72.65 &72.01 &71.35 &70.16 &68.59 \\
     25-phase &70.38 &68.37 &66.90 &65.33 &63.27 \\
    \specialrule{0.01em}{1.2pt}{1.7pt}
	\end{tabular}
	}
	\label{table:less_compressed_samples}
	\vspace{-.2cm}
\end{table*}

\end{document}